\documentclass[twoside]{article}
\usepackage{ecj,palatino,epsfig,latexsym,natbib}

\usepackage{hyperref}
\usepackage{graphicx}
\usepackage{xcolor}
\usepackage{multirow}
\usepackage{booktabs}
\usepackage{varwidth}
\usepackage{scrextend}
\usepackage{amsmath, dsfont}

\usepackage{amssymb}
\usepackage{colortbl}
\usepackage[counterclockwise, figuresright]{rotating}

\definecolor{mygrey}{RGB}{200, 200, 200}
\definecolor{mypink}{HTML}{E76BF3}

\definecolor{myred}{HTML}{F8766D}
\definecolor{mygreen}{HTML}{00BA38}
\definecolor{myblue}{HTML}{619CFF}

\definecolor{boxblue}{RGB}{0, 142, 151}
\definecolor{boxgreen}{RGB}{120, 183, 0}
\definecolor{boxred}{RGB}{153, 53, 73}
\definecolor{boxyellow}{RGB}{215, 175, 0}

\begin{document}

\title{\bf Automated Algorithm Selection on Continuous Black-Box Problems By Combining Exploratory Landscape Analysis and Machine Learning}  

\author{\name{\bf Pascal Kerschke} \hfill \addr{kerschke@uni-muenster.de}\\ 
        \addr{Information Systems and Statistics, University of M{\"u}nster, 48149 M{\"u}nster, Germany}
\AND
	\name{\bf Heike Trautmann} \hfill \addr{trautmann@uni-muenster.de}\\ 
        \addr{Information Systems and Statistics, University of M{\"u}nster, 48149 M{\"u}nster, Germany}
}

\maketitle

\begin{abstract}

In this paper, we build upon previous work on designing informative and efficient \textit{Exploratory Landscape Analysis} features for characterizing problems' landscapes and show their effectiveness in automatically constructing algorithm selection models in continuous black-box optimization problems.\newline
Focussing on algorithm performance results of the COCO platform of several years, we construct a representative set of high-performing complementary solvers and present an algorithm selection model that -- compared to the portfolio's single best solver -- on average requires less than half of the resources for solving a given problem. Therefore, there is a huge gain in efficiency compared to classical ensemble methods combined with an increased insight into problem characteristics and algorithm properties by using informative features.\newline
Acting on the assumption that the function set of the \textit{Black-Box Optimization Benchmark} is representative enough for practical applications the model allows for selecting the best suited optimization algorithm within the considered set for unseen problems \textit{prior} to the optimization itself based on a small sample of function evaluations. Note that such a sample can even be reused for the initial population of an evolutionary (optimization) algorithm so that even the feature costs become negligible.

\end{abstract}


\section{Introduction}
\label{sec:intro}

Although the \textit{Algorithm Selection Problem}~\citep[ASP,][]{Rice1975} has been introduced more than four decades ago, there only exist few works \citep[e.g.,][]{Bischl2012, Munoz2015_AS}, which perform algorithm selection in the field of continuous optimization. Independent of the underlying domain, the goal of the ASP can be described as follows: given a set of optimization algorithms $\mathcal{A}$, often denoted algorithm \textit{portfolio}, and a set of problem instances $\mathcal{I}$, one wants to find a model $m: \mathcal{I} \rightarrow \mathcal{A}$ that selects the best algorithm $A \in \mathcal{A}$ from the portfolio for an \textit{unseen} problem instance $I \in \mathcal{I}$. Albeit there already exists a plethora of optimization algorithms -- even when only considering single-objective, continuous optimization problems -- none of them can be considered to be superior to all the other ones across all optimization problems.  Hence, it is very desirable to find a sophisticated selection mechanism, which automatically picks the portfolio's best solver for a given problem. Of course hyper-heuristics \citep{Burke2003} already internally combining several algorithmic approaches can be part of the solver-portfolio as well.

\begin{figure}[!t]
\includegraphics[width=0.995\columnwidth, trim = 0mm 14mm 0mm 19mm, clip]{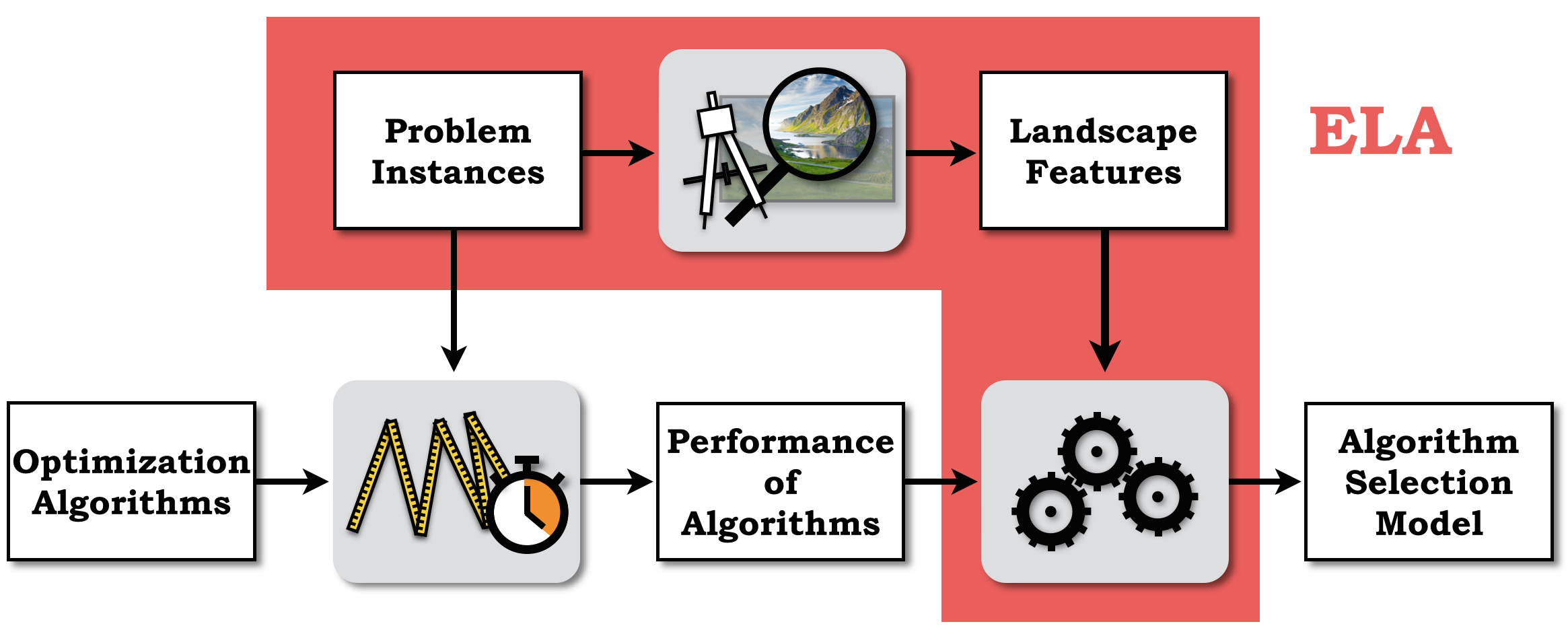}
\caption{Schematic view of how \textit{Exploratory Landscape Analysis} (ELA) can be used for improving the automated algorithm selection process.}
\label{fig:ela_as}
\end{figure}

Within other optimization domains, such as the well-known \textit{Travelling Salesperson Problem}, feature-based algorithm selectors have already shown their capability of outperforming the respective state-of-the-art optimization algorithm(s) by combining machine learning techniques and problem dependent features~\citep{Kotthoff2015, Kerschke2017_TSP}. As schematized in Figure~\ref{fig:ela_as}, we now transfer the respective idea of using instance-specific features to single-objective continuous optimization problems based on Exploratory Landscape Analysis \citep[ELA,][]{Mersmann2011} for leveraging  solver complementarity of a well-designed algorithm portfolio.

As we show within this work, the integration of ELA results in strong performance improvements (averaged across the entire benchmark) over \textit{any} of the portfolio's single solvers. More precisely, our selector requires only half of the resources needed by the portfolio's single best solver. Hence, our model strongly reduces the gap towards the idealistic -- and thus, from a realistic point of view unreachable -- virtual best solver. The latter is an oracle-like algorithm selector, which always predicts the best algorithm for a given problem, without using any additional information (such as problem features) and therefore without any additional costs.

To the best of our knowledge, this work is the first one of its kind in the domain of continuous black-box optimization, which successfully combines automatically computable features -- by means of ELA -- with sophisticated machine learning and feature selection techniques in order to model powerful algorithm selectors on a comprehensive, well-accepted algorithm portfolio and benchmark set. Our proposed approach is related to the field of \textit{automated machine learning} \citep[AutoML, see e.g.,][]{Feurer2015}, which tries to tackle machine learning problems (such as optimization and classification) in a completely automated fashion. Within our work, we extend the idea of automated machine learning with an additional assessment of the trained algorithm selection models and the analysis of the included ELA features. This allows for a better understanding of (a) the algorithm selectors, (b) the differences between the different optimization algorithms, and (c) the considered problem instances.

A more detailed overview of Exploratory Landscape Analysis, as well as an introduction into \texttt{flacco}~--~an extensive \texttt{R} toolbox for computing a variety of such landscape features  enhanced by a graphical user interface -- is given in Section~\ref{sec:ela}. In Section~\ref{sec:eda}, we give more insights into the COCO platform and afterwards describe our experimental setup, including the generation of the considered algorithm portfolio, in Section~\ref{sec:setup}. An analysis of our found algorithm selection models is given in Section~\ref{sec:results} and Section~\ref{sec:conclusion} concludes our work.

\section{Exploratory Landscape Analysis}
\label{sec:ela}

While problem-dependent (landscape) features can in general be computed for any optimization problem \citep[e.g.,][]{Mersmann2013,Hutter2014_algorithm,Ochoa2014,Pihera2014,Daolio2016}, we will only consider single-objective, continuous optimization problems within this work.

\begin{figure}[!t]
\centering
\includegraphics[width=\textwidth]{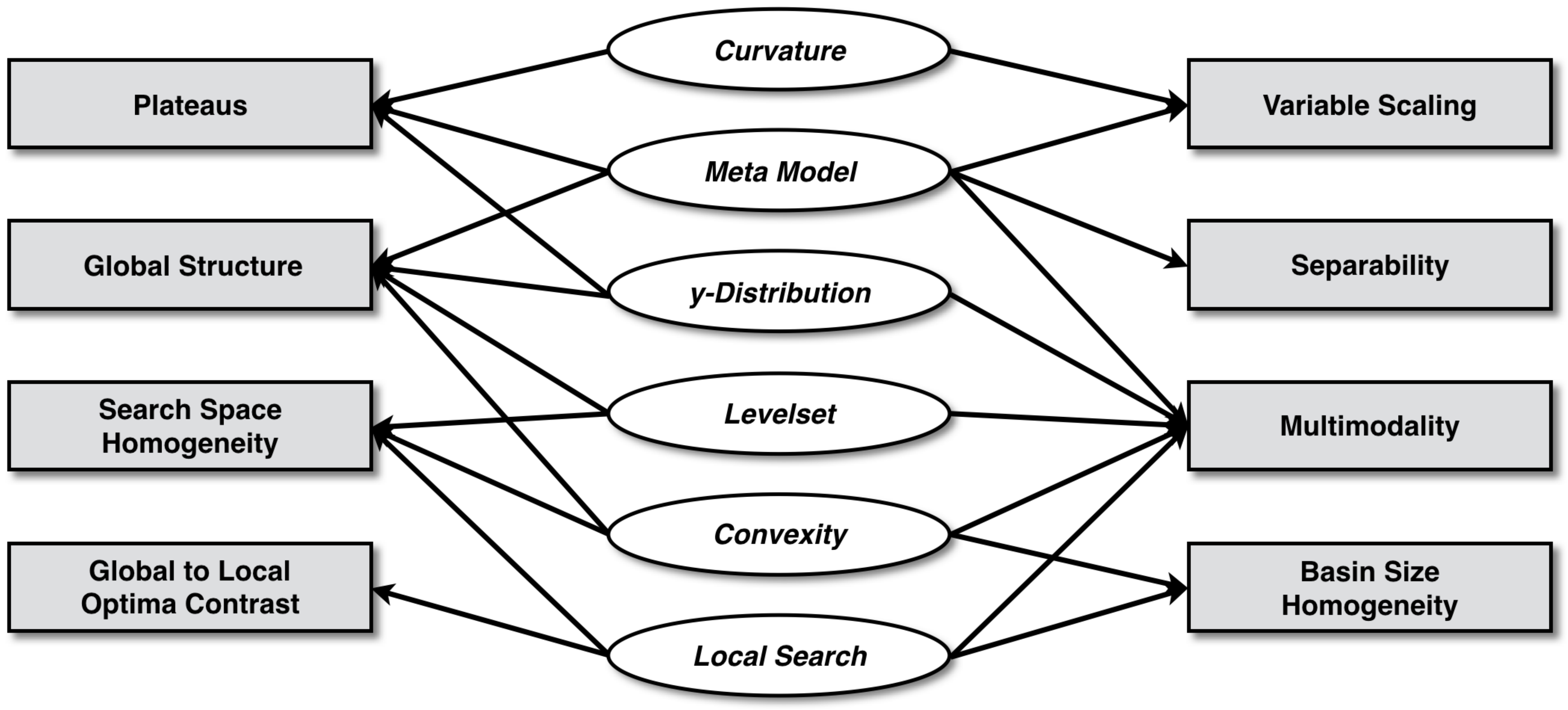}
\caption{Overview of relation between ``high-level'' properties (grey rectangles) and ``low-level'' features (white ellipses), taken from \cite{Mersmann2011}.}
\label{fig:ela}
\end{figure}

For this domain, \cite{Mersmann2011}~introduced a sophisticated approach for characterizing a problem's landscape by means of numerical feature values and called it \textit{``Exploratory Landscape Analysis''}. Within their work, they designed a total of 50 numerical measures and grouped them into six categories of so-called ``low-level'' features, which aim at numerically characterizing the landscape properties of a problem to be solved. The individual features are based on systematic sampling of the decision space by e.g., using a latin hypercube design. A wide range of mathematical and statistical measures, which can be calculated based on relatively few function evaluations, are included, and the respective composition of the features reflects the desired overall problem characteristics. Six low-level feature classes were introduced, i.e., measures related to the distribution of the objective function values ($y$-\textit{Distribution}), estimating meta-models such as linear or quadratic regression models on the sampled data (\textit{Meta-Model}) and the level of convexity (\textit{Convexity}). Furthermore, local searches are conducted starting at the initial design points (\textit{Local Search}), the relative position of each objective value compared to the median of all values is investigated (\textit{Levelset}), and numerical approximations of the gradient or the Hessian represent the class of curvature features (\textit{Curvature}). Each class comprises a set of sub-features which result from the same experimental data generated from the initial design. Figure~\ref{fig:ela} visualizes the assumed main relationships between the low-level feature classes and so-called ``high-level'' properties \citep{MersmannPTBW2015}, such as the degree of multimodality, the separability, the basin size homogeneity or the number of plateaus. However, given that these properties (a) require expert knowledge and as a consequence can not be computed automatically, and (b) are categorical and thus, make it for instance impossible to distinguish problems by their class of multimodality (none, low, medium, high), the introduction of the low-level features can be seen as a major step towards automatically computable landscape features and hence automated algorithm selection.

\subsection{Related Work}

Already in the years before the term ELA was introduced, researchers have tried to characterize a problem's landscape by numerical values: \cite{Jones1995fitness} assessed a problem's difficulty by means of a fitness distance correlation, \cite{Lunacek2006} introduced a dispersion metric, \cite{Malan2009} quantified the landscape's ruggedness and \cite{Mueller2011} performed fitness-distance analyses.

However, after \cite{Bischl2012} have shown the potential of landscape analysis by using ELA features for algorithm selection, a manifold of new landscape features has been introduced. \cite{Abell2013}~used hill climbing characteristics, \cite{Munoz2015_ELA}~measured a landscape's information content and \cite{Morgan2015} analyzed the problems with the help of length scale features. More recently, \cite{Malan2015}~characterized constrained optimization problems, and \cite{Shirakawa2016} introduced an entire bag of local landscape features.

In other works, research groups, which also include this paper's authors, have designed features based on a discretization of the continuous search space into a grid of cells~\citep{Kerschke2014}, and successfully employed the nearest better clustering approach for distinguishing funnel-shaped landscapes with a global structure from problems, whose local optima are aligned in a more random manner~\citep{Kerschke2015}. This particularly facilitates the decision of the class of optimization algorithms that suits best for the problem at hand. We also showed that even low budgets of $50 \times d$ observations ($d$ being the problem dimensionality) -- i.e.,~a sample size that is close to the size of an evolutionary algorithm's initial population -- is sufficient for such a distinction~\citep{Kerschke2016}. In consequence, the evaluation of this initial sample, which is required for the landscape features, would come without any additional costs, given that the evolutionary algorithm would have to evaluate those points anyway.

\subsection{Flacco}

In the previous subsection, we have provided an overview of numerous landscape features. Unfortunately,  those features were usually --~if available at all~-- implemented in different programming languages, such as Python~\citep{VanRossum2015}, Matlab~\citep{Matlab2013} or \texttt{R}~\citep{R2017}, making it extremely complicated to use all of them within a single experiment. This obstacle has been solved (for \texttt{R}-users) with the development of \texttt{flacco}~\citep{Kerschke2017_Flacco}, an \texttt{R}-package for \textbf{f}eature-based \textbf{l}andscape-\textbf{a}nalysis of \textbf{c}ontinuous and \textbf{c}onstrained \textbf{o}ptimization problems. The package (currently) provides a collection of more than 300 landscape features (including the majority of the ones from above), distributed across a total of 17 different feature sets. In addition, the package comes with several visualization techniques, which should help to facilitate the understanding of the underlying problem landscapes~\citep{Kerschke2016_CEC}. One can either use the package's stable release from CRAN\footnote{\url{https://cran.r-project.org/package=flacco}} or its developmental version from GitHub\footnote{\url{https://github.com/kerschke/flacco}}. Note that the latter also provides further information on the usage of \texttt{flacco}, as well as a link to its online-tutorial\footnote{\url{http://kerschke.github.io/flacco-tutorial/site/}}.

\begin{figure}[!t]
\centering
\includegraphics[width=0.7\textwidth, trim = 0mm 5mm 5mm 26mm, clip]{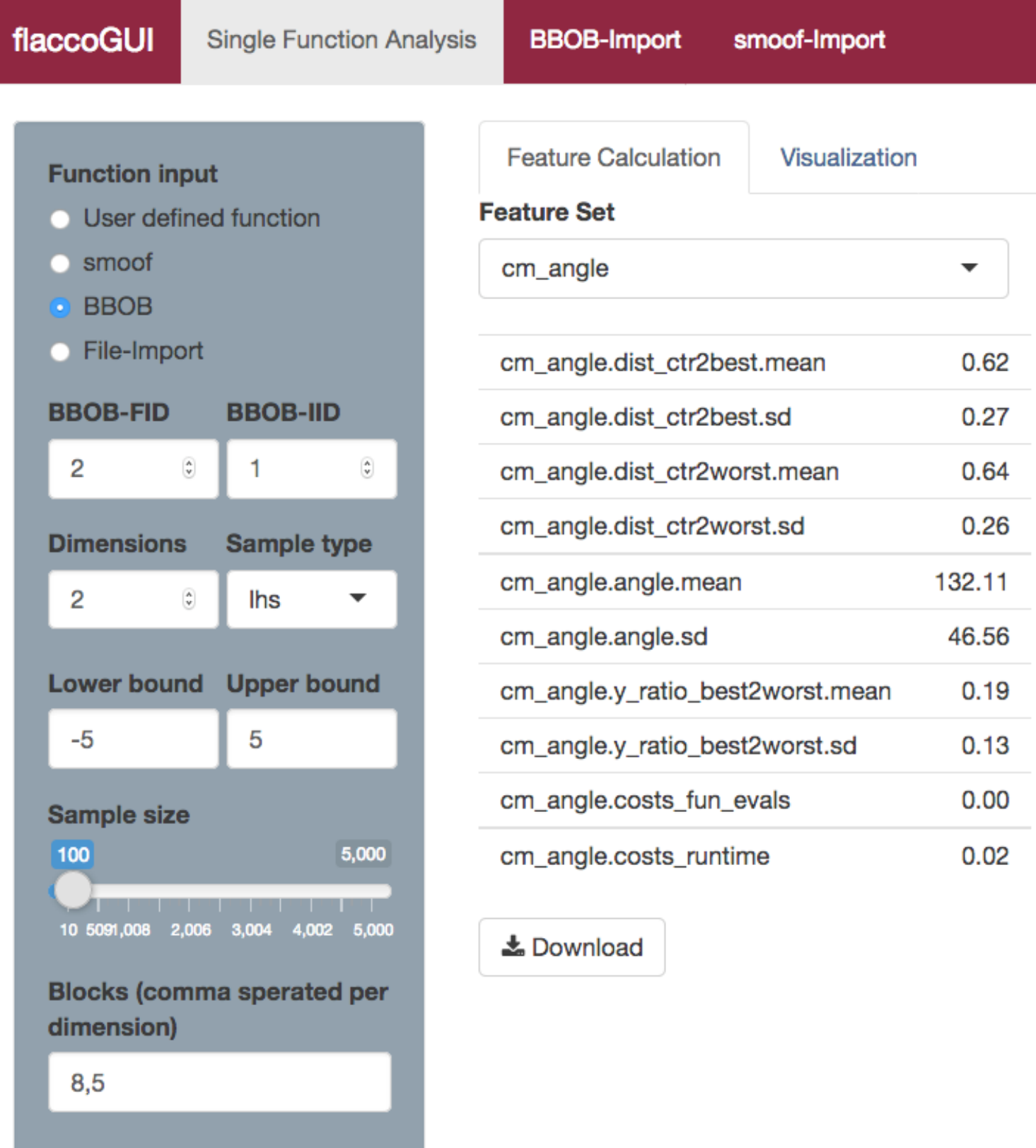}
\caption{Screenshot of the platform-independent GUI of \texttt{flacco}, which is hosted publicly available at \url{http://www.flacco.shinyapps.io/flacco}.}
\label{fig:flacco_gui}
\end{figure}

Being aware of possible obstacles for non-\texttt{R}-users, \cite{Hanster2017} recently developed a web-hosted and hence, platform-independent, graphical user interface (GUI)\footnote{\url{http://www.flacco.shinyapps.io/flacco}} of \texttt{flacco}. A screenshot of that GUI is displayed in Figure~\ref{fig:flacco_gui}. The GUI provides a slim, and thus user-friendly, version of \texttt{flacco}. In its left panel (highlighted in grey), the user needs to provide information on the function that should be optimized -- either by selecting one of the single-objective optimization problems available in the \texttt{R}-package \texttt{smoof}~\citep{Bossek2016}, configuring one of the BBOB-functions, manually defining a function, or by uploading an evaluated set of points. On the right side, one then can either compute (and download) a specific feature set or visualize certain aspects of the optimization problem and/or its features.

\section{Exploratory Data Analysis of COCO}
\label{sec:eda}

Instead of executing the optimization algorithms ourselves, we use the results from COCO\footnote{\url{http://coco.gforge.inria.fr/}}~\citep{Hansen2016}, which is a platform for \textbf{CO}mparing \textbf{C}ontinuous \textbf{O}ptimization algorithms on the Black-Box Optimization Benchmark \citep[BBOB,][]{Hansen2009, Finck2010}. The platform provides a collection of the performance results of 129 optimization algorithms, which have been submitted to the BBOB workshops at the GECCO conference\footnote{\url{http://sig.sigevo.org/index.html/tiki-index.php?page=GECCOs}} of the years 2009, 2010, 2012, 2013 and 2015. Using the results published on this platform comes with the advantage that over the years basically all important classes of optimization algorithms, including current state-of-the art optimizers, have been submitted. Thus, experiments are of representative character in a comprehensive setting.

\subsection{General Setup of the COCO-Platform}

The competition settings were quite similar across the years: per dimension $d~\in~\{2, 3, 5, 10, 20, 40\}$ and function $\text{FID}~\in~\{1, \ldots, 24\}$, the participants had to submit the results for a total of 15 problem instances. As shown below, only five instances (IIDs~1~to~5) were part of \textit{each} competition, while the remaining ten instances changed per year:
\begin{itemize}
\item 2009: IIDs 1 to 5 (with 3 replications each)
\item 2010: IIDs 1 to 15 (1 replication each)
\item 2012: IIDs 1 to 5, 21 to 30 (1 replication each)
\item 2013: IIDs 1 to 5, 31 to 40 (1 replication each)
\item 2015: IIDs 1 to 5, 41 to 50 (1 replication each)
\end{itemize}
For each pair of problem instance $i \in \mathcal{I}$ and optimization algorithm $A \in \mathcal{A}$, the submitted data contains a log of the performed number of function evaluations and the corresponding achieved fitness value, enabling an a posteriori evaluation of the solver's performance. More precisely, this data allows to decide whether the solver was \textit{successful} to approximate the (known) global optimum of instance $i \in \mathcal{I}$ up to a precision value $\varepsilon~\in~\{10^1, 10^0, \ldots, 10^{-7}\}$ and also, how many function evaluations $FE_i (\varepsilon)$ were performed until the solver (un)successfully terminated. Here, a solver is \textit{successful} w.r.t. the precision threshold $\varepsilon$, i.e., $Success_i(\varepsilon) = 1$, if it found a solution $x^{\ast} \in [-5, +5]^d$, whose fitness value $f(x^{\ast})$ lies within $[f(x_{opt}), f(x_{opt}) + \varepsilon]$. Note that this definition implies that the found solution $x^{\ast}$ does not necessarily have to be located in the (decision space's) close neighborhood of the global optimum $x_{opt}$ -- as \textit{success} is exclusively defined via the proximity in the objective space. Then, $Success_i (\varepsilon)$ and $FE_i (\varepsilon)$ were used to compute the solver's \textit{Expected Runtime}~\citep[ERT,][]{Hansen2009}:
\[ERT (\varepsilon) = \frac{\sum_i FE_i (\varepsilon)}{\sum_i Success_i (\varepsilon)}.\]

Although the aforementioned setup comes with some drawbacks, we will still use it as it has been well-established in the community and thus allows for wide comparability. Nevertheless, we would like to address its pitfalls as well: (a) Given the strong variability across the instances' objective values, it is not straightforward to use a fixed \textit{absolute} precision value for comparing the solvers across all instances. Instead, it might be more reasonable to use a relative threshold. (b)~Computing the ERT based on different instances -- even if they belong to the same BBOB problem -- is very questionable as the instances can have completely different landscapes: if the original instance is a highly multimodal function, the transformed instance\footnote{Per definition, instances are identical up to rotation, scaling and shifts~\citep{Finck2010}.} could -- if at all -- consist of only a few peaks. Therefore, we strongly encourage to ask for several solver runs on the same instance in future setups, i.e., perform (at least five to ten) replicated runs, and then evaluate the performance results per instance rather than per function allowing for ERT computations on function \textit{and} instance level.

\subsection{Preprocessing COCO's Solver Performances}

We will use the performance results of the 129 optimization algorithms available at COCO\footnote{An overview of all submitted optimization algorithms along with their descriptions can be found at \url{http://coco.gforge.inria.fr/doku.php?id=algorithms}.}. However, in order to work with a valid data base, we first performed some sanity checks on the platform's data before combining it in a joint data base. For this purpose, we first combined the algorithms' performance results in a joint data base and afterwards check whether they fulfill each competition's requirements regarding the required number of runs on the correct instances.

While the submitted results of all 29 (2009) and 23 (2010) solvers of the first two competitions passed these checks, in the following years, only 22 of 28 (2012), 16 of 31 (2013) and 13 of 18 (2015) submissions did so. The invalid submissions \textit{partially} used settings of the previous years.
However, in order to use the most general portfolio of solvers, we only considered the submissions for IIDs~1~to~5 with only one run per instance, as this is the only set of instances that was used across all five BBOB competitions. Fortunately, the performances of all 129 solvers could be considered for our experiments, because even the problematic submissions from above had valid data for the first five instances.

\section{Experimental Setup}
\label{sec:setup}

\subsection{Algorithm Performance Data}
\label{sec:exp_perf}
For each of the 129 solvers we computed the ERT per tuple of problem dimension $d \in \{2, 3, 5, 10\}$, BBOB problem respectively function $\text{FID} \in \{1, \ldots, 24\}$ and problem instance $\text{IID} \in \{1, \ldots, 5\}$ (if multiple runs exist, we only considered the first run), resulting in a total of $61\,920$ observations. The ERTs were computed for a precision threshold of $\varepsilon = 10^{-2}$, because smaller values led to too many unsuccessful runs. Even for this chosen precision, only approximately $67\%$ of all (considered) runs terminated successfully. For better comparability across the different problems, a standardized version of the ERT (the \textit{relative ERT}) will be used (see Section~\ref{sec:perf}).


\subsection{Instance Feature Data}
\label{sec:exp_feats}
Each of the ELA feature sets was computed using a so-called \textit{improved latin hypercube design}~\citep{Beachkofski2002} consisting of $50 \times d$ observations, which were sampled across the decision space, i.e., $[-5, +5]^d$. The feature sets were then computed using the \texttt{R}-package \texttt{flacco}~\citep{Kerschke2017_Flacco} for all four problem dimensions, 24 BBOB problems and five problem instances that were used by the performance data (see Section~\ref{sec:exp_perf}). For each of these 480 problems, we calculated the six `classical' ELA feature sets from \cite{Mersmann2011} (convexity, curvature, levelset, local search, meta-model and y-distribution), as well as the basic, (cell mapping) angle\footnote{The cell mapping angle features were computed using three blocks per dimension, as larger values would result in too many empty cells due to the ``curse of dimensionality''.}~\citep{Kerschke2014}, dispersion~\citep{Lunacek2006}, information content~\citep{Munoz2015_ELA}, nearest better clustering~\citep{Kerschke2015} and principal component features, resulting in a total of 102 features per problem instance.

Although being conscious of the resulting information loss, we aggregate each feature across the five problem instances (per BBOB problem) via the median of the respective feature values, in order to map our feature data to the 96 observations (24 problems, four dimensions) of the performance data.


\subsection{Constructing the Algorithm Portfolio}
\label{sec:exp_portfolio}

For meaningfully addressing the algorithm selection task, the choice of the underlying algorithm portfolio is crucial. Ideally, the considered set should be as small and as complementary as possible and should include state-of-the art optimizers.  For this purpose, we ranked the solvers per considered BBOB problem based on ERT performance. We then constructed four solver sets (one per dimension), each of them containing the solvers that ranked within the ``Top~3'' of at least one of the 24 functions of the respective dimension. Based on these four solver sets -- i.e., one per considered problem dimension, and each of them consisting of 37 to 41 solvers -- a portfolio of 12 solvers was constructed by only considering optimizers that belonged to \textit{each} of the four sets.

The 12 optimization algorithms from the found portfolio can be grouped into four categories and are summarized below.

\subsubsection{Deterministic Optimization Algorithms (2)}

The two solvers of this category are variants of the Brent-STEP algorithm\footnote{The Brent-STEP algorithm itself accelerates the global line search method STEP \citep[``select the easiest point'',][]{Swarzberg1994} by using Brent's method~\citep{Brent2013}.}~\citep{Baudivs2015}. It performs axis-parallel searches and chooses the next iteration's search dimension either using a round-robin~\citep[\textbf{BSrr},][]{Povsik2015} or a quadratic interpolation strategy~\citep[\textbf{BSqi},][]{Povsik2015}.

\subsubsection{Multi-Level Approaches (5)}

The origin of most solvers belonging to this category is the \textit{multi level single linkage} method~\citep[\textbf{MLSL},][]{Pal2013, Rinnooy1987}. It is a stochastic, multistart, global optimizer that relies on random sampling and local searches. Aside from MLSL itself, some of its variants also belong to our portfolio: an interior-point version for constrained nonlinear problems~\citep[\textbf{fmincon},][]{Pal2013}, a quasi-Newton version, which approximates the Hessian using BFGS~\citep{Broyden1970} \citep[\textbf{fminunc},][]{Pal2013}, and a hybrid variant whose most important improvements are related to its sampling phase \citep[\textbf{HMLSL},][]{Pal2013}. The final optimizer belonging to this group is the \textit{multilevel coordinate search} \citep[\textbf{MCS},][]{Huyer2009}, which splits the search space into smaller boxes -- each containing a known observation -- and then starts local searches from promising boxes.

\subsubsection{Variants of the CMA-ES (4)}

\cite{Hansen2001} introduced one of the most popular evolution strategies: the \textit{Covariance Matrix Adaption Evolution Strategy (CMA-ES)} with cumulative step-size adaptation \citep[\textbf{CMA-CSA},][]{Atamna2015}. It led to a plethora of variants~\citep{VanRijn2016}, including the following three solvers from our portfolio: (1)~\textbf{IPOP400D}~\citep{Auger2013}, a restart version of the CMA-ES with an increasing population size \citep[IPOP-CMA-ES,][]{Auger2005} and a maximum of $400 \times (d + 2)$ function evaluations. (2)~A~hybrid CMA \citep[\textbf{HCMA},][]{Loshchilov2013}, which combines a \textit{bi-population self-adaptive surrogate-assisted CMA-ES}\footnote{A BIPOP-CMA-ES \citep{Hansen2009_BIPOP} is a multistart CMA-ES with equal budgets for two interlaced restart strategies: one with an increasing population size and one with varying small population sizes.} \citep[BIPOP-$^{s\ast}$aACM-ES-k,][]{Loshchilov2013intensive}, STEP~\citep{Swarzberg1994} and NEWUOA~\citep{Powell2006} to benefit from surrogate models and line searches simultaneously. (3) A \textit{sequential, model-based algorithm configuration} \citep[SMAC,][]{Hutter2011} procedure applied to the BBOB problems \citep[\textbf{SMAC-BBOB},][]{Hutter2013}. It uses Gaussian processes \citep[GP,][]{Rasmussen2006} to model the the \textit{expected improvement function} and then performs one run of DIRECT~\citep{Jones1993} (with $10\times d$ evaluations) and ten runs of the classical CMA-ES \citep{Hansen2001} (with $100\times d$ evaluations) on the expected improvement function.

\subsubsection{Others (1)}

The final optimizer from our portfolio is called OptQuest/NLP \citep[\textbf{OQNLP},][]{Pal2013,Ugray2007}. It is a commercial, heuristic, multistart algorithm that was designed to find the global optima of smooth constrained nonlinear programs (NLPs) and mixed integer nonlinear programs (MINLPs). The algorithm uses the \textit{OptQuest Callable Library} \citep[OCL,][]{Laguna2003} to generate candidate starting points for a local NLP solver.


\subsection{Machine Learning Algorithms}
\label{sec:ml}

We considered three classes of supervised learning strategies for training our algorithm selection models: (1)~A \textit{classification} approach, which simply tries to predict the best-performing optimizer\footnote{Ties are broken via random uniform sampling among the tied solvers.} and hence, completely ignores the magnitude of performance differences between the best and the remaining portfolio solvers. (2)~A \textit{regression} approach, which trains a separate model for the performances of each optimizer and afterwards predicts the solver with the best predicted performance. (3) In addition to these well-known strategies, we also considered the so-called \textit{pairwise regression}, which led to promising results in other works \citep[e.g.,][]{Kotthoff2015,Kerschke2017_TSP}. In contrast to modeling the performances straightforwardly (as in (2)), it models the performance differences for each solver pair and afterwards predicts the solver whose predicted performance difference was the highest, compared to all other solvers.

The algorithm selectors were trained in \texttt{R}~\citep{R2017} using the \texttt{R}-package \texttt{mlr}~\citep{Bischl2016_mlr}. For each class of the considered supervised learning approaches (i.e., classification, regression and paired regression), we used recursive partitioning and regression trees \citep[\texttt{rpart},][]{rpart}, kernel-based support vector machines \citep[\texttt{ksvm},][]{ksvm}, random forests \citep[\texttt{randomForest},][]{randomforest} and extreme gradient boosting \citep[\texttt{xgboost},][]{xgboost}. Additionally, we also tried multivariate adaptive regression splines \citep[\texttt{mars},][]{mars} in case of the (paired) regression approaches.

Note that the SVM's inverse kernel width \texttt{sigma} was the only hyperparameter that was (automatically) configured -- using the \texttt{sigest} function from the \texttt{R}-package \texttt{kernlab}~\citep{ksvm}. All other hyperparameters were used in their default settings: the SVMs used Gaussian radial basis kernels and the random forests were constructed using 500 trees, whose split points were sampled random uniformly of $\lfloor \sqrt{p} \rfloor$ (classification) or $\max \{\lfloor p / 3 \rfloor, \; 1\}$ (regression / paired regression) features with $p$ being the data set's number of (ELA) features.


\subsection{Feature Selection Strategies}
\label{sec:featsel}

Each of the aforementioned 14 algorithm selectors (see Section~\ref{sec:ml}), i.e., 4 using a classification-based approach, 5 using regression and 5 using paired regression, are initially trained using the set of all 102 features (see Section~\ref{sec:exp_feats}). However, using all of the features simultaneously likely causes lots of noise and/or redundancy, which could lead to poorly performing algorithm selectors. Furthermore, some of the feature sets, namely, the convexity, curvature and local search features from the classical ELA features~\citep{Mersmann2011}, require additional function evaluations on top of the costs for the initial design. In order to overcome these obstacles, we used the following four feature selection strategies -- all of them are implemented in \texttt{mlr} -- to train further algorithm selectors. The resulting 56 additional models (four feature selection strategies times 14 machine learning algorithms) consist of smaller and likely much more meaningful feature subsets, which on the one hand lead to better performing selectors and on the other hand improve the interpretability of the trained models.

\begin{itemize}
\item \textit{Greedy forward-backward selection (sffs):} This feature selection strategy starts with an empty feature set and iteratively alternates between greedily adding and/or removing features as long as the model's performance improves.

\item \textit{Greedy backward-forward selection (sfbs):} Analogously to sffs, this approach greedily adds and removes features per iteration. However, in contrast to the previous method, sfbs starts with the full set of all 102 features and afterwards alternates between removing and adding features.

\item \textit{$(10+5)$-GA:} A genetic algorithm~\citep[GA, see e.g.,][]{Eiben2015} using \texttt{mlr}'s default values for the population ($\mu = 10$) and offspring ($\lambda = 5$) sizes is used. For this approach, the selected features are represented as a 102-dimensional bit string, where a  value of 1 at position $k$ implies that the $k$-th feature is selected. The GA runs for a maximum of 100 generations and selects the features by performing random bit flips -- with a (default) mutation rate of $5\%$ and a crossover rate of $50\%$.

\item \textit{$(10+50)$-GA:} A slightly modified version of the previous GA, which uses the tenfold of offsprings (50 instead of 5) per generation in order to increase the selection pressure within the GA's evolutionary loop.
\end{itemize}


\subsection{Performance Assessment}
\label{sec:perf}

In lieu of using the ERT itself, we will use the \textit{relative ERT (relERT)} for our experiments. While the former strongly biases the algorithm selectors towards multimodal and higher-dimensional problems due to much larger amounts of used function evaluations, the relERTs, which are also used within the BBOB competitions, allow a fair comparison of the solvers across the problems and dimensions by normalizing each solver's ERT with the ERT of the best solver for the respective BBOB problem (of that dimension).
Instead of scaling each performance with the respective best ERT of all 129 solvers, we used the best performance from the 12 solvers of the considered portfolio as this is our set of interest in this study. 

As some solvers did not even solve a single instance from some of the BBOB functions, the respective ERTs and relERTs were not defined. We thus imputed the missing relERTs using the well-known PAR10 score \citep[e.g.,][]{Bischl2016}. That is, each of the problematic values is replaced with a penalty value ($36\,690.3$) that is the tenfold of the highest valid relERT; for all other values, the respective (finite) relERTs are used.

For each of the supervised learning approaches (see Section~\ref{sec:ml}), the algorithm selectors are evaluated using the \textit{mean relative ERT}, which averages the relERTs of the predictions (including the costs for the initial design) on the corresponding test data, i.e., a subset of the BBOB problems. In order to obtain more realistic and reliable estimates of the selectors' performances, they were assessed using leave-one-(function)-out cross-validation. That is, per algorithm selector we train a total of 96 submodels (24 BBOB problems in four problem dimensions each). Each of them uses only 95 of the 96 BBOB problems for training and the remaining one for testing. Note that each problem was used exactly once as test data. As a consequence, within each iteration/fold of the leave-one-(function)-out cross-validation, exactly one problem (i.e., the respective test data) is completely kept out of the modeling phase and only used for assessing the respective submodel's performance. The average of the resulting 96 relERTs is then used as our algorithm selector's performance.

Following common practices in algorithm selection \citep[e.g.,][]{Bischl2016}, we compare the performances of our algorithm selectors with two baselines: the \textit{virtual best solver} (VBS) and the \textit{single best solver} (SBS). The virtual best solver, sometimes also called \textit{oracle} or \textit{perfect selector}, provides a lower bound for the selectors as it shows the performance that one \textit{could} (theoretically) achieve on the data, when always selecting the best performing algorithm per instance. Given that the relERT has a lower bound of 1 and that at least one solver achieves that perfect performance per instance, the VBS has to be 1 per definition. Nevertheless, it is quite obvious that algorithm selectors usually do not reach such an idealistic performance given their usually imperfect selections and/or the influence of the costs for computing the landscape features.

The second baseline, the SBS, represents the (aggregated) performance of the (single) best solver from the portfolio. Consequently, this baseline is much more important than the VBS, because an algorithm selector is only useful if its performance (including feature costs) is better than the SBS. Note that in principle, the SBS could either be determined on the entire data set or, for (leave-one-out) cross-validation, be computed per fold. However, within our experiments, both approaches result in identical performances.


\subsection{Overview of the Interlinks Between our Framework's Building Blocks}
\label{sec:overview}

\begin{figure}[!t]
\includegraphics[width = \textwidth, trim = 0mm 20mm 0mm 2mm, clip]{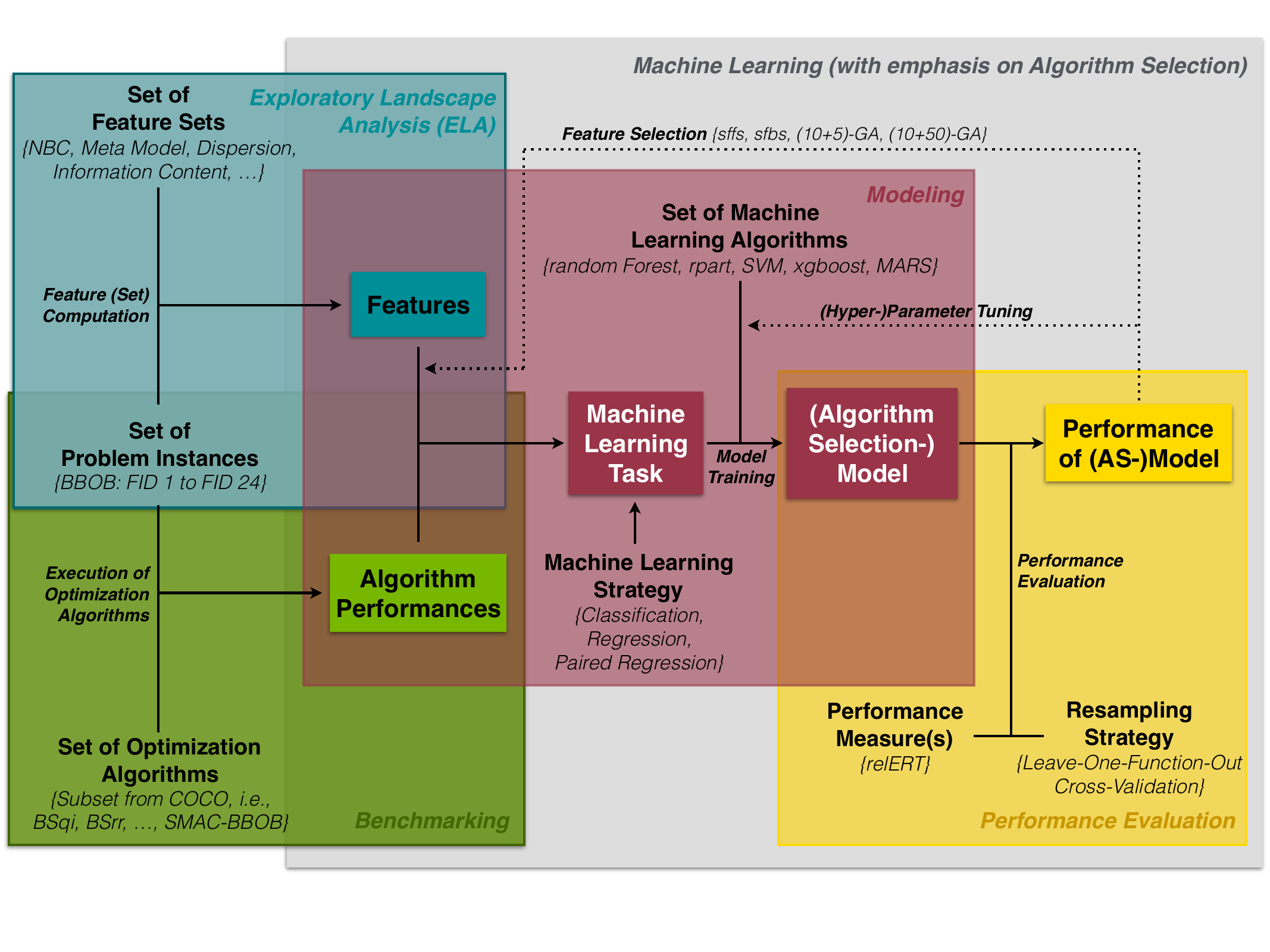}
\caption{This scheme is a slightly modified version of a figure taken from \cite{Kerschke2017_Diss}, showing the interlinks between \textit{feature-based problem characterization} (ELA; \textcolor{boxblue}{blue} area in the top left), \textit{benchmarked performances} of the considered optimization algorithms (\textcolor{boxgreen}{green} area in the bottom left), the actual \textit{modeling} process of the machine learning algorithms (\textcolor{boxred}{red} area in the center) and the \textit{evaluation} of the algorithm selectors' performances (\textcolor{boxyellow}{yellow} area in the bottom right). The grey box in the background reveals the connections of the aforementioned topics to the field of \textit{machine learning} in general.}
\label{fig:ml_scheme}
\end{figure}

In the previous Sections, we have outlined the separate building blocks for our considered algorithm selection framework. The connections and therefore the interactions between these blocks are displayed in Figure~\ref{fig:ml_scheme}.

The benchmarked algorithm performances (Sections~\ref{sec:exp_perf} and \ref{sec:exp_portfolio}; visualized by means of a \textcolor{boxgreen}{green} box in the bottom left of Figure~\ref{fig:ml_scheme}) and the corresponding automatically computable and problem-specific landscape features (Section~\ref{sec:exp_feats}; \textcolor{boxblue}{blue} box in the top left) are used by different machine learning algorithms (Section~\ref{sec:ml}; \textcolor{boxred}{red} box in the center) for training several possible algorithm selection models. Their performances are then assessed (Section~\ref{sec:perf}; \textcolor{boxyellow}{yellow} in the bottom right) and in turn used as feedback mechanism to improve the previously trained algorithm selection models, e.g., by selecting more informative feature subsets (Section~\ref{sec:featsel}).

As shown within the grey area of Figure~\ref{fig:ml_scheme}, further feedback mechanisms, such as tuning the considered machine learning techniques, are also an option and in general very promising. However, in this first study of its kind (within the domain of single-objective continuous optimization), we rely on the default configurations of the machine learning algorithms (except for one very important parameter of the SVM as detailed in Section~\ref{sec:ml}), which already provides lots of interesting and promising results -- as will be described in Section~\ref{sec:results}. The quite complex issue of optimally configuring the machine learning algorithms will be addressed in future work.

\section{Results}
\label{sec:results}

\subsection{Analyzing the Algorithm Portfolio}

\begin{figure}[!t]
\centering
\includegraphics[width=0.8\textwidth]{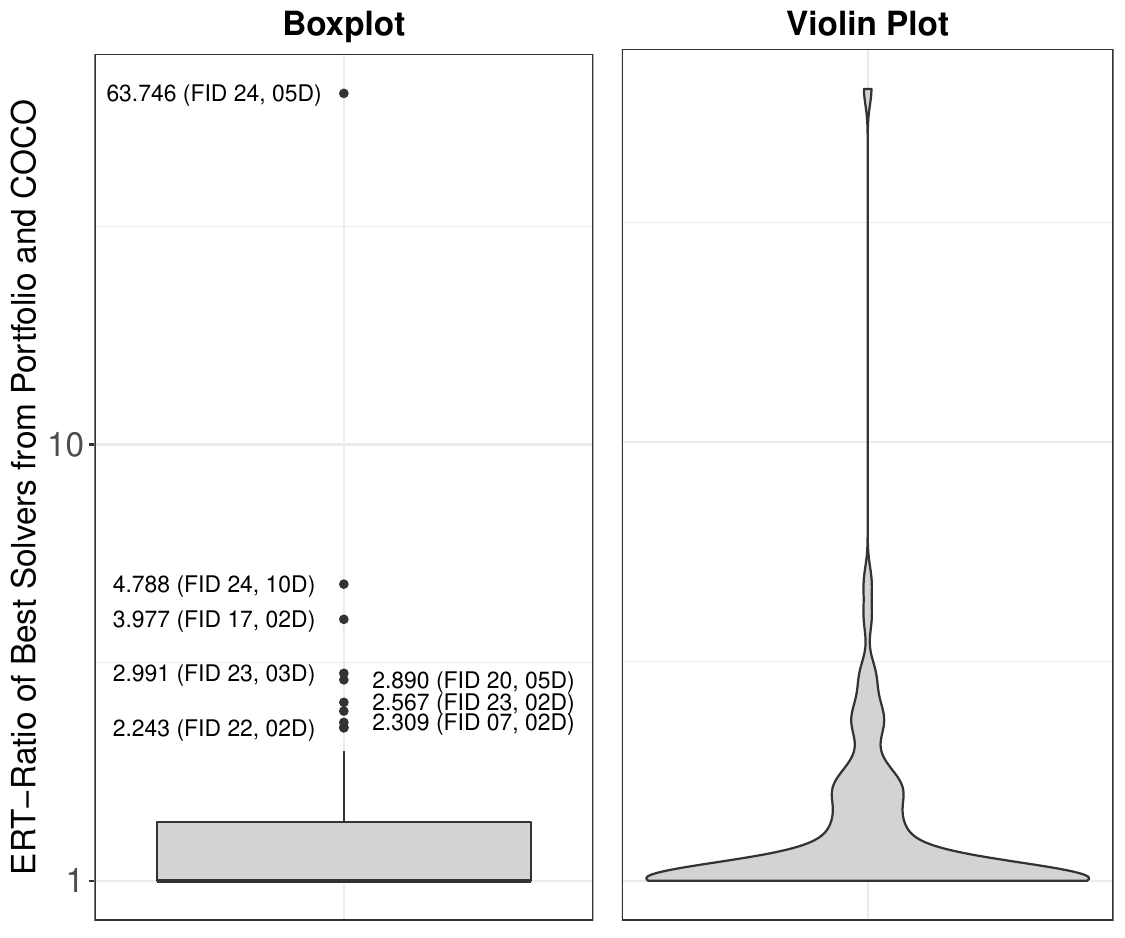}
\caption{Boxplot (left) and violin plot (right) visualizing the ratios (on a log-scale) between the ERTs of the portfolio's best solvers and the best ones from COCO. The labeled outliers within the left image, coincidentally depicting all problems with a ratio greater than two, indicate the BBOB problems that were difficult to solve for the 12 algorithms from the considered portfolio. Note that two instances, FID 18 in 10D (ratio:~2.251) and FID 20 in 10D (ratio:~2.451), have not been labeled within the plot to account for better readability.}
\label{fig:boxplot}
\end{figure}

Here, we examine the constructed portfolio to get more insights into the data provided by COCO. In a first step, we compare the best performances of the portfolio solvers to the best performances of all 129 solvers. For 51 of the 96 considered BBOB functions over the dimensions, the portfolio's VBS is identical to the VBS of all COCO-solvers.

The violin plot on the right side of Figure~\ref{fig:boxplot} -- which can be understood as a boxplot, whose shape represents a kernel density estimation of the underlying observations -- indicates that the ratios between the best ERTs from the portfolio and all COCO-solvers (per BBOB problem) is usually close to one. More precisely, only ten BBOB problems, depicted by black dots in the boxplot of Figure~\ref{fig:boxplot}, have a ratio greater than two, and only three of them -- the 2D-version of \textit{Schaffers F7 Function} \citep[FID 17,][]{Finck2010}, as well as the 5D- and 10D-versions of the \textit{Lunacek bi-Rastrigin Function} \citep[FID 24,][]{Finck2010} -- have ratios worse than three. Consequently, we can conclude that \textit{if} we find a well-performing algorithm selector, its suggested optimization algorithm likely requires less than twice the number of function evaluations compared to the best algorithm from the entire COCO platform.

As mentioned in Section~\ref{sec:perf}, we had to impute some of the performances, because not all optimizers solved at least one instance per BBOB problem. More precisely, HCMA was the only one to find at least one valid solution for all 96 problems, followed by HMLSL (88), CMA-CSA (87), OQNLP (72), fmincon (71), MLSL (69), fminunc (68), IPOP400D and MCS (67), BSqi (52), BSrr (50) and SMAC-BBOB (18). Hence, the mean relERT of HCMA (30.37) was by far the lowest of all considered solvers, making this solver the clear SBS.

However, as shown in Table~\ref{tab:portfolio}, HCMA is not superior to the remaining solvers across all problems. Independent of the problem dimensions, the Brent-STEP variants BSqi and BSrr outperform the other optimizers on the separable problems (FIDs 1 to 5). Similarly, the multilevel methods MLSL, fmincon and HMLSL are superior on the unimodal functions with high conditioning (FIDs 10 to 14), and the multimodal problems with adequate global structure (FIDs 15 to 19) are solved fastest by MCS (in 2D) and CMA-CSA (3D to 10D). The remaining problems, i.e., functions with low or moderate conditioning (FIDs 6 to 9), as well as multimodal functions with a weak global structure (FIDs 20 to 24) do not show such obvious patterns. Nevertheless one can see that HCMA, HMLSL, OQNLP and sometimes also MLSL perform quite promising on these function groups.
\begin{sidewaystable}[p]
\caption{Summary of the relative expected runtime of the 12 portfolio solvers and the two algorithm selectors (including costs for feature computation) shown per dimension and BBOB group. The single best solver (SBS) for each of these combinations is printed in \textbf{bold} and the overall SBS (HCMA) is highlighted in grey. Values in \textbf{\textit{bold italic}} indicate that the respective algorithm selector was better than any of the single portfolio solvers. Further details on this summary are given within the text.}
\label{tab:portfolio}
\centering
\scriptsize
\begin{tabular}{cc|rrrrr>{\columncolor[gray]{0.8}}rrrrrrr|rr}
  \toprule  & \multicolumn{1}{c}{} & \multicolumn{14}{c}{Relative Expected Runtime of the 12 Solvers from the Considered Algorithm Portfolio and the 2 Presented Algorithm Selection-Models}\\
  \cmidrule{3-16} \multirow{-2}{*}{Dim} & \multirow{-3}{*}{BBOB-} & \multirow{2}{*}{BSqi} & \multirow{2}{*}{BSrr} & CMA- & \multirow{2}{*}{fmincon} & \multirow{2}{*}{fminunc} & & \multirow{2}{*}{HMLSL} & IPOP- & \multirow{2}{*}{MCS} & \multirow{2}{*}{MLSL} & \multirow{2}{*}{OQNLP} & SMAC- & \multicolumn{2}{c}{AS-Model}\\ 
  & \multirow{-3}{*}{Group} & & & CSA &  &  & \multirow{-2}{*}{HCMA} & & 400D & & & & BBOB & \# 1 & \# 2 \\ 
  \midrule \multirow{6}{*}{2} & F1 - F5 & \textbf{1.2} & 1.3 & 54.8 & 11.0 & 11.8 & 3.7 & 14.6 & 18.4 & 5.8 & 15.5 & 17.0 & \hspace*{-0.2cm}22\,014.9 & 16.6 & 20.3 \\ 
   & F6 - F9 & 18\,516.7 & \hspace*{-0.2cm}9\,708.2 & 7.4 & 18.6 & 19.2 & 5.8 & 1.7 & 5.7 & 11.3 & 24.2 & \textbf{1.5} & \hspace*{-0.2cm}27\,518.6 & 3.1 & 3.5 \\ 
   & F10 - F14 & 7\,649.2 & \hspace*{-0.2cm}7\,481.5 & 8.3 & \textbf{1.0} & 62.7 & 6.3 & \textbf{1.0} & 10.7 & 322.7 & \textbf{1.0} & 4.9 & \hspace*{-0.2cm}29\,353.2 & 4.7 & 4.0 \\ 
   & F15 - F19 & 7\,406.6 & \hspace*{-0.2cm}14\,710.3 & 14.7 & \hspace*{-0.2cm}7\,392.0 & \hspace*{-0.2cm}7\,367.7 & 25.3 & 8.1 & 15.5 & \textbf{7.7} & \hspace*{-0.2cm}7\,391.7 & \hspace*{-0.2cm}7\,351.2 & \hspace*{-0.2cm}29\,354.8 & 26.2 & 10.1 \\ 
   & F20 - F24 & 84.8 & \hspace*{-0.2cm}14\,768.5 & \hspace*{-0.2cm}7\,351.9 & 4.1 & 14.5 & 44.9 & 3.9 & \hspace*{-0.2cm}14\,679.3 & 11.4 & \textbf{2.1} & 2.7 & \hspace*{-0.2cm}22\,014.6 & 42.5 & 3.0 \\ 
   \cmidrule{2-16} & all & 6\,240.7 & \hspace*{-0.2cm}9\,318.4 & \hspace*{-0.2cm}1\,549.1 & \hspace*{-0.2cm}1\,546.5 & \hspace*{-0.2cm}1\,556.7 & 17.7 & \textbf{6.0} & \hspace*{-0.2cm}3\,068.4 & 74.3 & \hspace*{-0.2cm}1\,547.9 & \hspace*{-0.2cm}1\,536.9 & \hspace*{-0.2cm}25\,990.1 & 19.3 & 8.4 \\ 
   \midrule \multirow{6}{*}{3} & F1 - F5 & \textbf{1.3} & 1.3 & \hspace*{-0.2cm}7\,367.9 & 85.2 & 132.1 & 356.1 & 6.8 & \hspace*{-0.2cm}14\,686.6 & 45.9 & 55.9 & \hspace*{-0.2cm}7\,347.6 & \hspace*{-0.2cm}22\,015.1 & 58.4 & 94.9 \\ 
   & F6 - F9 & 331.2 & \hspace*{-0.2cm}9\,527.4 & 4.7 & 38.5 & \hspace*{-0.2cm}9\,173.7 & 4.5 & \textbf{1.9} & 6.5 & 31.4 & \hspace*{-0.2cm}9\,173.4 & 2.5 & \hspace*{-0.2cm}36\,690.3 & 3.3 & 39.9 \\ 
   & F10 - F14 & 29\,356.3 & \hspace*{-0.2cm}14\,712.1 & 8.9 & \textbf{1.0} & 4.1 & 5.0 & 1.0 & 12.3 & \hspace*{-0.2cm}8\,132.7 & 1.0 & 9.3 & \hspace*{-0.2cm}29\,353.4 & 4.8 & 3.6 \\ 
   & F15 - F19 & 14\,698.2 & \hspace*{-0.2cm}22\,026.2 & \textbf{1.6} & \hspace*{-0.2cm}14\,701.2 & \hspace*{-0.2cm}14\,699.5 & 2.6 & 11.4 & \hspace*{-0.2cm}7\,339.4 & \hspace*{-0.2cm}7\,346.9 & \hspace*{-0.2cm}14\,700.0 & \hspace*{-0.2cm}14\,686.2 & \hspace*{-0.2cm}36\,690.3 & 2.8 & 7.1 \\ 
   & F20 - F24 & 14\,741.8 & \hspace*{-0.2cm}14\,758.7 & \hspace*{-0.2cm}7\,389.4 & \hspace*{-0.2cm}7\,339.6 & \hspace*{-0.2cm}14\,677.4 & 66.8 & 2.3 & \hspace*{-0.2cm}22\,015.1 & \hspace*{-0.2cm}7\,342.4 & \hspace*{-0.2cm}7\,339.8 & \textbf{1.9} & \hspace*{-0.2cm}22\,014.8 & 67.0 & 3.4 \\ 
   \cmidrule{2-16} & all & 12\,304.7 & \hspace*{-0.2cm}12\,316.7 & \hspace*{-0.2cm}3\,077.4 & \hspace*{-0.2cm}4\,616.2 & \hspace*{-0.2cm}7\,677.5 & 90.4 & \textbf{4.8} & \hspace*{-0.2cm}9\,178.9 & \hspace*{-0.2cm}4\,769.4 & \hspace*{-0.2cm}6\,132.4 & \hspace*{-0.2cm}4\,593.1 & \hspace*{-0.2cm}29\,047.1 & 28.3 & 29.4 \\ 
   \midrule \multirow{6}{*}{5} & F1 - F5 & \textbf{1.4} & 1.4 & \hspace*{-0.2cm}7\,533.6 & \hspace*{-0.2cm}14\,678.4 & \hspace*{-0.2cm}14\,679.2 & 12.0 & 17.5 & \hspace*{-0.2cm}14\,688.7 & \hspace*{-0.2cm}14\,678.1 & \hspace*{-0.2cm}14\,678.5 & \hspace*{-0.2cm}14\,678.0 & \hspace*{-0.2cm}22\,015.1 & 22.7 & 22.9 \\ 
   & F6 - F9 & 27\,597.4 & \hspace*{-0.2cm}36\,690.3 & 5.6 & \hspace*{-0.2cm}9\,173.5 & \hspace*{-0.2cm}9\,173.8 & 3.9 & \textbf{2.4} & 4.9 & 28.8 & \hspace*{-0.2cm}9\,173.4 & \hspace*{-0.2cm}9\,173.5 & \hspace*{-0.2cm}36\,690.3 & 4.8 & 4.8 \\ 
   & F10 - F14 & 22\,032.8 & \hspace*{-0.2cm}29\,360.3 & 8.9 & 1.0 & 11.9 & 4.2 & \textbf{1.0} & 13.6 & \hspace*{-0.2cm}22\,019.2 & \textbf{1.0} & 10.7 & \hspace*{-0.2cm}36\,690.3 & 5.2 & 5.2 \\ 
   & F15 - F19 & 36\,690.3 & \hspace*{-0.2cm}36\,690.3 & \textbf{3.1} & \hspace*{-0.2cm}36\,690.3 & \hspace*{-0.2cm}36\,690.3 & 4.3 & \hspace*{-0.2cm}7\,346.1 & \hspace*{-0.2cm}29\,352.5 & \hspace*{-0.2cm}36\,690.3 & \hspace*{-0.2cm}36\,690.3 & \hspace*{-0.2cm}29\,352.5 & \hspace*{-0.2cm}36\,690.3 & 4.4 & 4.4 \\ 
   & F20 - F24 & 22\,053.6 & \hspace*{-0.2cm}22\,050.8 & \hspace*{-0.2cm}7\,400.0 & \hspace*{-0.2cm}14\,678.9 & \hspace*{-0.2cm}22\,014.9 & \textbf{7.7} & \hspace*{-0.2cm}7\,339.8 & \hspace*{-0.2cm}22\,017.4 & \hspace*{-0.2cm}14\,681.0 & \hspace*{-0.2cm}22\,015.0 & \hspace*{-0.2cm}14\,676.8 & \hspace*{-0.2cm}22\,014.9 & 7.8 & 7.8 \\ 
   \cmidrule{2-16} & all & 21\,428.3 & \hspace*{-0.2cm}24\,469.8 & \hspace*{-0.2cm}3\,114.6 & \hspace*{-0.2cm}15\,289.0 & \hspace*{-0.2cm}16\,819.9 & \textbf{6.5} & \hspace*{-0.2cm}3\,063.8 & \hspace*{-0.2cm}13\,765.8 & \hspace*{-0.2cm}18\,352.4 & \hspace*{-0.2cm}16\,817.4 & \hspace*{-0.2cm}13\,761.8 & \hspace*{-0.2cm}30\,575.6 & 9.1 & 9.2 \\ 
   \midrule \multirow{6}{*}{10} & F1 - F5 & \textbf{1.6} & 1.6 & \hspace*{-0.2cm}14\,691.0 & \hspace*{-0.2cm}14\,679.9 & \hspace*{-0.2cm}14\,682.7 & 2.7 & \hspace*{-0.2cm}7\,365.5 & \hspace*{-0.2cm}14\,698.8 & \hspace*{-0.2cm}14\,680.0 & \hspace*{-0.2cm}14\,679.9 & \hspace*{-0.2cm}14\,678.3 & \hspace*{-0.2cm}22\,015.7 & 16.3 & 16.3 \\ 
   & F6 - F9 & 36\,690.3 & \hspace*{-0.2cm}27\,563.9 & 4.3 & \hspace*{-0.2cm}9\,173.4 & \hspace*{-0.2cm}9\,173.8 & \textbf{2.2} & 4.1 & \hspace*{-0.2cm}9\,181.9 & \hspace*{-0.2cm}9\,188.1 & \hspace*{-0.2cm}9\,173.4 & \hspace*{-0.2cm}9\,173.9 & \hspace*{-0.2cm}36\,690.3 & 2.7 & 2.7 \\ 
   & F10 - F14 & 29\,359.3 & \hspace*{-0.2cm}29\,359.8 & 8.4 & 1.1 & 15.4 & 2.8 & \textbf{1.1} & \hspace*{-0.2cm}7\,352.5 & \hspace*{-0.2cm}22\,018.7 & \textbf{1.1} & 12.0 & \hspace*{-0.2cm}36\,690.3 & 3.7 & 3.7 \\ 
   & F15 - F19 & 36\,690.3 & \hspace*{-0.2cm}36\,690.3 & \textbf{1.7} & \hspace*{-0.2cm}36\,690.3 & \hspace*{-0.2cm}36\,690.3 & 2.0 & \hspace*{-0.2cm}22\,028.5 & \hspace*{-0.2cm}29\,352.5 & \hspace*{-0.2cm}36\,690.3 & \hspace*{-0.2cm}36\,690.3 & \hspace*{-0.2cm}36\,690.3 & \hspace*{-0.2cm}36\,690.3 & 2.1 & 2.1 \\ 
   & F20 - F24 & 36\,690.3 & \hspace*{-0.2cm}29\,367.0 & \hspace*{-0.2cm}14\,685.9 & \hspace*{-0.2cm}22\,015.2 & \hspace*{-0.2cm}22\,015.0 & \textbf{23.6} & \hspace*{-0.2cm}14\,677.1 & \hspace*{-0.2cm}29\,352.8 & \hspace*{-0.2cm}22\,018.9 & \hspace*{-0.2cm}22\,014.6 & \hspace*{-0.2cm}22\,014.9 & \hspace*{-0.2cm}36\,690.3 & 23.7 & 23.7 \\ 
   \cmidrule{2-16} & all & 27\,519.5 & \hspace*{-0.2cm}24\,472.9 & \hspace*{-0.2cm}6\,123.0 & \hspace*{-0.2cm}16\,817.8 & \hspace*{-0.2cm}16\,821.3 & \textbf{6.9} & \hspace*{-0.2cm}9\,182.4 & \hspace*{-0.2cm}18\,354.6 & \hspace*{-0.2cm}21\,408.0 & \hspace*{-0.2cm}16\,817.6 & \hspace*{-0.2cm}16\,819.7 & \hspace*{-0.2cm}33\,633.1 & 10.0 & 10.0 \\ 
   \midrule \multirow{6}{*}{all} & F1 - F5 & \textbf{1.4} & 1.4 & \hspace*{-0.2cm}7\,411.8 & \hspace*{-0.2cm}7\,363.6 & \hspace*{-0.2cm}7\,376.5 & 93.6 & \hspace*{-0.2cm}1\,851.1 & \hspace*{-0.2cm}11\,023.1 & \hspace*{-0.2cm}7\,352.4 & \hspace*{-0.2cm}7\,357.4 & \hspace*{-0.2cm}9\,180.2 & \hspace*{-0.2cm}22\,015.2 & 28.5 & 38.6 \\ 
   & F6 - F9 & 20\,783.9 & \hspace*{-0.2cm}20\,872.4 & 5.5 & \hspace*{-0.2cm}4\,601.0 & \hspace*{-0.2cm}6\,885.1 & 4.1 & \textbf{2.5} & \hspace*{-0.2cm}2\,299.8 & \hspace*{-0.2cm}2\,314.9 & \hspace*{-0.2cm}6\,886.1 & \hspace*{-0.2cm}4\,587.9 & \hspace*{-0.2cm}34\,397.4 & 3.5 & 12.7 \\ 
   & F10 - F14 & 22\,099.4 & \hspace*{-0.2cm}20\,228.4 & 8.7 & 1.0 & 23.5 & 4.6 & \textbf{1.0} & \hspace*{-0.2cm}1\,847.3 & \hspace*{-0.2cm}13\,123.3 & \textbf{1.0} & 9.3 & \hspace*{-0.2cm}33\,021.8 & 4.6 & 4.1 \\ 
   & F15 - F19 & 23\,871.3 & \hspace*{-0.2cm}27\,529.3 & \textbf{5.2} & \hspace*{-0.2cm}23\,868.5 & \hspace*{-0.2cm}23\,861.9 & 8.6 & \hspace*{-0.2cm}7\,348.5 & \hspace*{-0.2cm}16\,515.0 & \hspace*{-0.2cm}20\,183.8 & \hspace*{-0.2cm}23\,868.1 & \hspace*{-0.2cm}22\,020.0 & \hspace*{-0.2cm}34\,856.4 & 8.9 & 5.9 \\ 
   & F20 - F24 & 18\,392.6 & \hspace*{-0.2cm}20\,236.3 & \hspace*{-0.2cm}9\,206.8 & \hspace*{-0.2cm}11\,009.4 & \hspace*{-0.2cm}14\,680.5 & \textbf{35.8} & \hspace*{-0.2cm}5\,505.8 & \hspace*{-0.2cm}22\,016.2 & \hspace*{-0.2cm}11\,013.4 & \hspace*{-0.2cm}12\,842.9 & \hspace*{-0.2cm}9\,174.1 & \hspace*{-0.2cm}25\,683.7 & 35.3 & \textit{\textbf{9.5}} \\ 
   \cmidrule{2-16} & all & 16\,873.3 & \hspace*{-0.2cm}17\,644.5 & \hspace*{-0.2cm}3\,466.0 & \hspace*{-0.2cm}9\,567.4 & \hspace*{-0.2cm}10\,718.9 & \textbf{30.4} & \hspace*{-0.2cm}3\,064.3 & \hspace*{-0.2cm}11\,091.9 & \hspace*{-0.2cm}11\,151.0 & \hspace*{-0.2cm}10\,328.8 & \hspace*{-0.2cm}9\,177.9 & \hspace*{-0.2cm}29\,811.5 & \textit{\textbf{16.7}} & \textit{\textbf{14.2}} \\ 
   \bottomrule \end{tabular}
\end{sidewaystable}

Interestingly, several solvers had extremely high PAR10-scores for multiple BBOB-groups. The most obvious one is SMAC-BBOB, whose relative ERT (aggregated across the instances of a BBOB group) was higher than 10\,000 on all (!) BBOB groups. However, this finding is plausible: although this optimization algorithm belongs to the ``Top~3'' of at least one problem per dimension (according to the setup of the portfolio, see Section~\ref{sec:exp_portfolio}) and therefore to the ``Top~3'' of at least four problems across the entire benchmark, it only managed to solve 18 of the considered 96 problem instances (up to the required precision of $\varepsilon = 10^{-2}$). This in turn means that it failed on more than 80\% of all instances and thus the corresponding performances were set to the penalty value ($36\,690.3$, see Section~\ref{sec:perf}). Based on the fact that Table~\ref{tab:portfolio} only displays performances that were aggregated across at least four problems each, SMAC-BBOB's scores look that extreme -- and consequently might be misleading to a certain extent. 

\begin{figure}[!t]
\includegraphics[width=\textwidth]{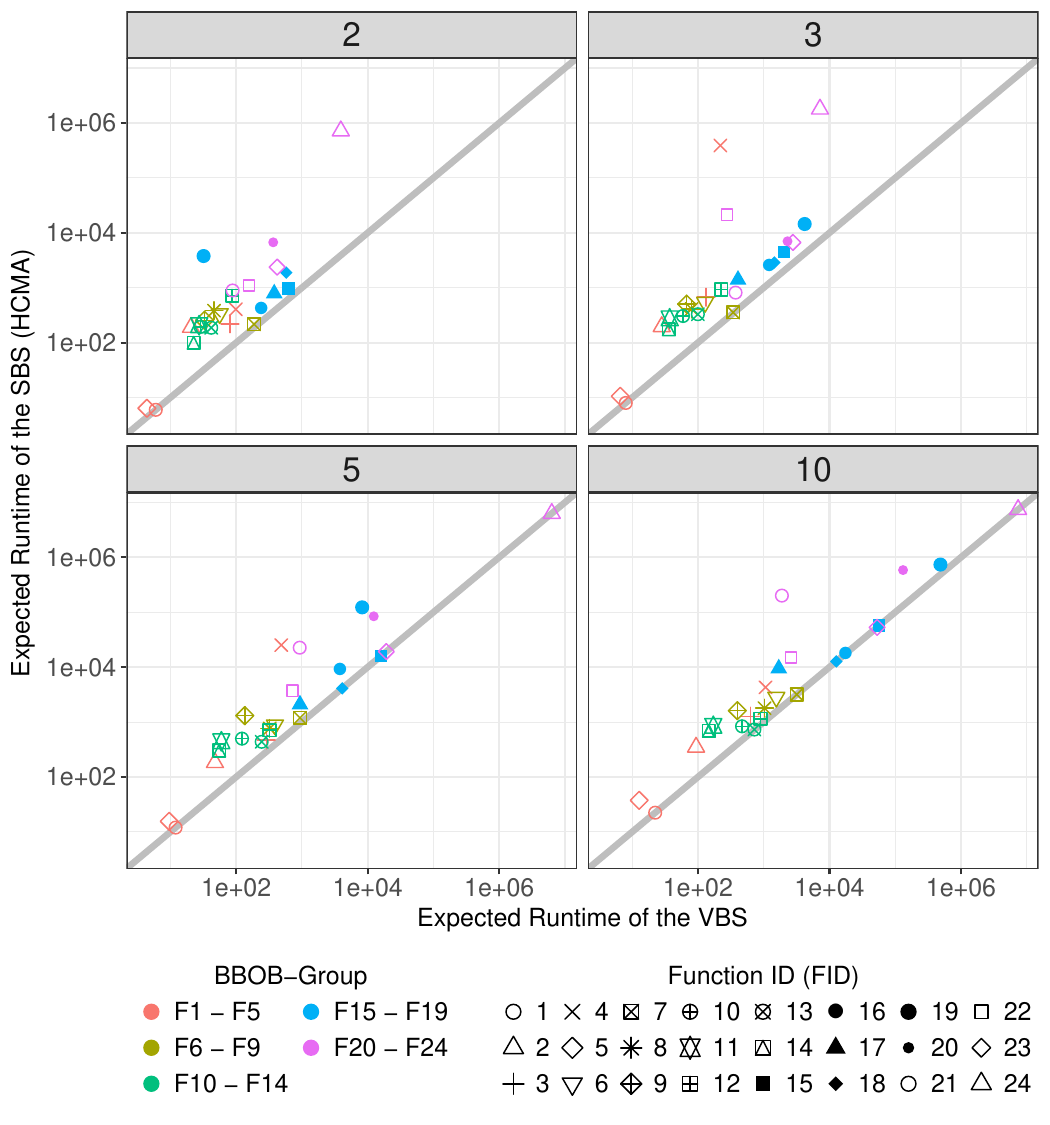}
\caption{Scatterplots comparing the \textit{expected runtime (ERT)} (shown on a log-scale) of the virtual best solver (VBS) and the single best solver (SBS), i.e., HCMA, across the 24 BBOB functions per problem dimension: 2D (top left), 3D (top right), 5D (bottom left) and 10D (bottom right).}
\label{fig:sbs_vbs}
\end{figure}

A further finding of Table~\ref{tab:portfolio} is that HCMA often is inferior to the other 11 solvers. This clearly indicates that the data contains sufficient potential for improving over the SBS -- e.g., with a reasonable algorithm selector. This is supported by the small amount of points located exactly on the diagonal of the scatterplots in Figure~\ref{fig:sbs_vbs}, which depict the ERT-pairs of the two baseline algorithms -- the idealistic VBS and the SBS, i.e., the best performing solver from the portfolio (HCMA) -- for all 24 BBOB problems and per problem dimension. Noticeably, though not very surprising, the VBS and SBS always have higher ERTs on the multimodal problems (FIDs 15 to 24) than on the unimodal problems (FIDs~1~to~14). Also, the shape of the cloud of observations indicates that the problems' complexity grows with its dimension: while the 2D-observations are mainly clustered in the lower left corner (indicating rather easy problems), the cloud stretches along the entire diagonal for higher-dimensional problems. Furthermore it is obvious that two problems, FIDs~1 (\textit{Sphere}, indicated by~\textcolor{red}{$\circ$}) and 5 (\textit{Linear Slope}, \textcolor{red}{$\diamond$}), are consistently solved very fast (independent of its dimensionality), whereas FID~24 (\textit{Lunacek Bi-Rastrigin}, \textcolor{mypink}{$\triangle$}) is always the most difficult problem for the two baseline algorithms.

Looking at the performances of HCMA across all 96 problems, one problem stands out: the three-dimensional version of the \textit{B\"uche-Rastrigin} function (FID 4). On this (usually rather simple) function, HCMA has a relative ERT of 1\,765.9, compared to at most 51.8 on the non-3D-versions of this problem and 249.1 as the worst relative ERT of all remaining 95 BBOB problems. We therefore had a closer look at the submitted data of HCMA across FID 4 and for all dimensions. The respective data is summarized in Table~\ref{tab:fid4}. Comparing the results of the 20 optimization runs with each other, three of them seem to be clear outliers. However, it remains unclear whether they were caused by the algorithm itself -- it could for instance be stuck in a local optimum without terminating -- or whether the outliers were caused by some technical issue on the platform.

\begin{table}[t]
\caption{Comparison of the number of function evaluations required by the portfolio's single best solver, i.e.,~HCMA, for solving the 20 instances of FID~4 (\textit{B\"uche-Rastrigin}) up to a precision of $10^{-2}$.}
\label{tab:fid4}
\centering
\begin{tabular}{rc|cr|c|cc}
\toprule
\hspace{-0.1cm}\multirow{2}{*}{Dim}\hspace{-0.1cm}	&	\hspace{-0.1cm}\multirow{2}{*}{IID}\hspace{-0.1cm}	&	Gap to		&	\hspace{-0.18cm}\# Function	& 	\hspace{-0.18cm}\multirow{2}{*}{Success?}\hspace{-0.18cm}  & \multicolumn{1}{c}{ERT of VBS} & \multicolumn{1}{c}{ERT of SBS}\\
   			&		&	Optimum		&	\hspace{-0.18cm}Evaluations	&   & \multicolumn{1}{c}{(Solver)} & \multicolumn{1}{c}{(relERT)} \\ \midrule
	&	1	&	8.44e-03    &	215			&	TRUE  & & \\
	&	2	&	6.91e-03    &	545			&	TRUE  & \multirow{2}{*}{98.8}& \multirow{2}{*}{405.2} \\
2	&	3	&	8.84e-05    &	440			&	TRUE  & \multirow{2}{*}{(BSqi)} & \multirow{2}{*}{(4.1)} \\
	&	4	&	5.44e-03    &	248			&	TRUE  & & \\
	&	5	&	4.29e-03    &	578			&	TRUE  & & \\ \midrule
	&	1	&	4.48e-03    &	\hspace{-0.18cm}976\,690		&	TRUE  & & \\
	&	2	&	9.99e-01    &	\hspace{-0.18cm}570\,925		&	FALSE  & \multirow{2}{*}{219.6} & \multirow{2}{*}{387\,784.5} \\
3	&	3	&	9.62e-04    &	781			&	TRUE  & \multirow{2}{*}{(BSrr)} & \multirow{2}{*}{(1\,765.9)} \\
	&	4	&	1.72e-03    &	1\,373		&	TRUE  & & \\
	&	5	&	8.24e-03    &	1\,369		&	TRUE  & & \\ \midrule
	&	1	&	6.52e-03    &	2\,048		&	TRUE  & & \\
	&	2	&	8.09e-04    &	2\,248		&	TRUE  & \multirow{2}{*}{486.2} & \multirow{2}{*}{25\,197.0} \\
5	&	3	&	1.00e+00    &	\hspace{-0.18cm}91\,882		&	FALSE  & \hspace{-0.18cm}\multirow{2}{*}{(BSrr)} & \multirow{2}{*}{(51.8)} \\
	&	4	&	2.16e-03    &	2\,382		&	TRUE  & & \\
	&	5	&	8.54e-03    &	2\,228		&	TRUE  & & \\ \midrule
	&	1	&	3.75e-03    &	4\,253		&	TRUE  & & \\
	&	2	&	6.72e-03    &	4\,495		&	TRUE  & \multirow{2}{*}{1\,067.8} & \multirow{2}{*}{4\,286.6} \\
10	&	3	&	2.58e-03    &	4\,632		&	TRUE  & \multirow{2}{*}{(BSrr)} & \multirow{2}{*}{(4.0)} \\
	&	4	&	2.79e-03    &	4\,032		&	TRUE  & & \\
	&	5	&	5.43e-03    &	4\,021		&	TRUE  & & \\ \bottomrule
\end{tabular}
\end{table}


\subsection{Analyzing the Best Algorithm Selector}
\label{sec:model1}

After gaining more insights into the underlying data, we use the performance and feature data for training a total of 70 algorithm selectors: 14 machine learning algorithms (see Section~\ref{sec:ml}), each of them using either one of the four feature selection strategies described in Section~\ref{sec:featsel} or no feature selection at all. The classification-based support vector machine, which employed a greedy forward-backward feature selection strategy (sffs), achieved the best performance. While the single best solver (HCMA) had a mean relative ERT of 30.37, the algorithm selector (denoted as ``Model 1'') had a mean relative ERT of 16.67 (including feature costs) and 13.60 (excluding feature costs). Therefore, the selector is able to pick an optimizer from the portfolio, which solves the respective problem on average twice as fast.

Eight features resulted from the feature selection approach and were included in Model 1: three features from the y-distribution feature set \citep[the skewness, kurtosis and number of peaks of a kernel-density estimation of the problems' objective values; see][]{Mersmann2011}, one levelset feature \citep[the ratio of mean misclassification errors when using a linear (LDA) and mixed discriminant analysis (MDA);][]{Mersmann2011}, two information content features \citep[the \textit{maximum information content} and the \textit{settling sensitivity}; see][]{Munoz2015_ELA}, one cell mapping feature \citep[the standard deviation of the distances between each cell's center and worst observation; see][]{Kerschke2014} and one of the basic features (the best fitness value within the sample).

\begin{figure}[!p]
\includegraphics[width=0.985\textwidth]{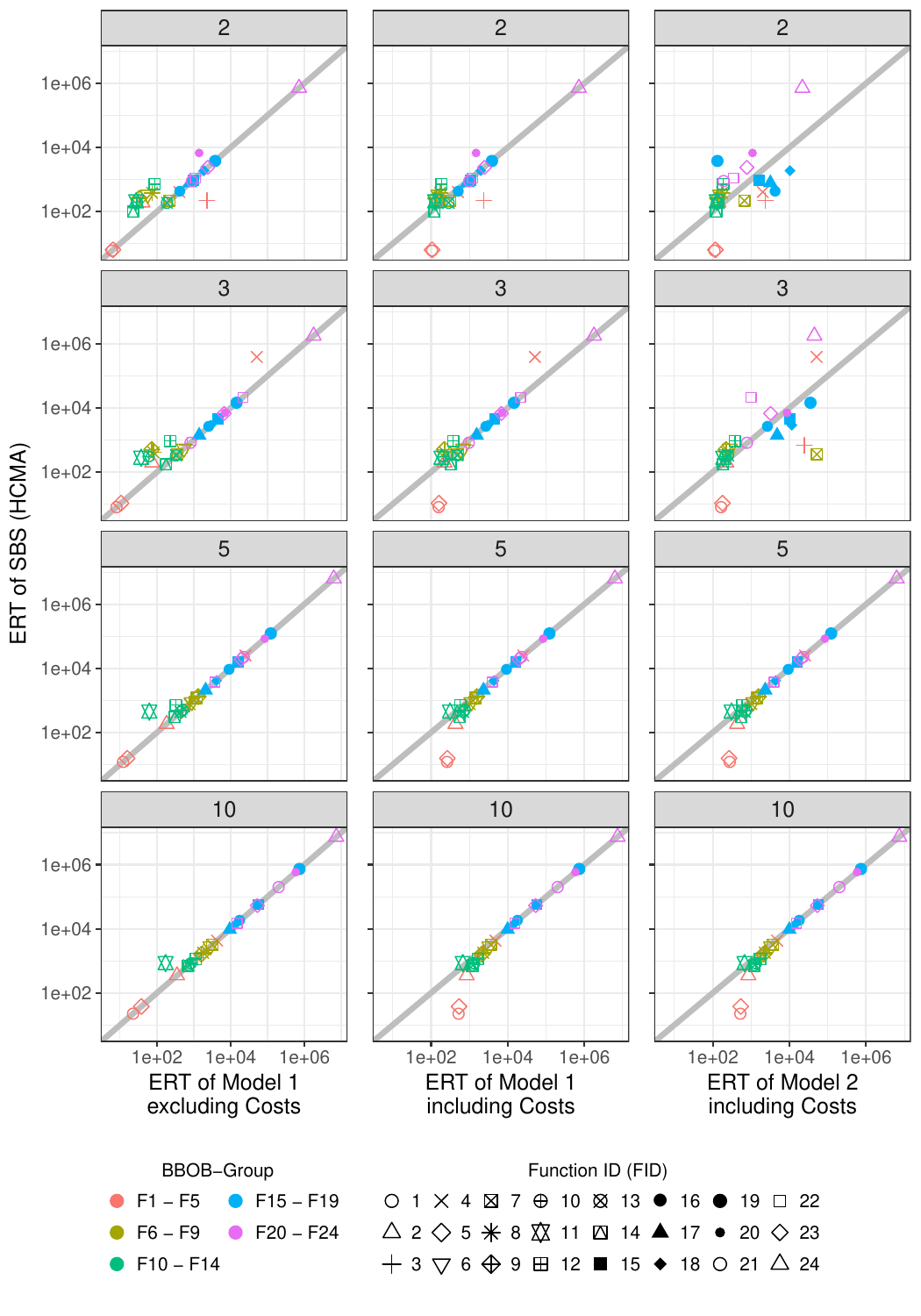}
\caption{Scatterplots of the ERT-pairs (shown on a log-scale) of the SBS (i.e., HCMA) and Model 1 without considering the costs for computing the features (left column), Model 1 including the feature costs (middle), and Model 2 including the feature costs (right).}
\label{fig:sbs_mod_all}
\end{figure}

Looking at the scatterplots shown in the left column of Figure~\ref{fig:sbs_mod_all}, one can see that -- with the exception of the 2D \textit{Rastrigin} function (FID 3, \textcolor{myred}{+}) -- Model 1 predicts on all instances an optimizer that performs at least as good as HCMA. Admittedly, this comparison is not quite fair, as the ERTs shown within the respective column do not consider the costs for computing the landscape features. Hence, a more realistic comparison of the selector and the SBS is shown in the second column of Figure~\ref{fig:sbs_mod_all}.

The fact that Model~1 performs worse than HCMA on some of the problems is negligible as the losses only occur on rather simple -- and thus, quickly solvable -- problems. The allocated budget for computing the landscape features was $50 \times d$, i.e., between 100 and 500 function evaluations depending on the problem dimension. On the other hand, the ERT of the portfolio's VBS lies between 4.4 and 7\,371\,411. These numbers (a) explain, why Model 1 performs poorly on rather simple problems, and (b) support the thesis that the costs for computing the landscape features only have a small impact on the total costs when solving rather complex problems. However, one should be aware of the fact that the number of required function evaluations for the initial design is in a range of the common size of initial (evolutionary) algorithm populations so that -- given that the majority of (promising) optimization algorithms belongs to the group of evolutionary algorithms -- in practice no additional feature costs would occur.

\begin{figure}[!t]
\includegraphics[width=\textwidth, trim = 1mm 1mm 0mm 0mm, clip]{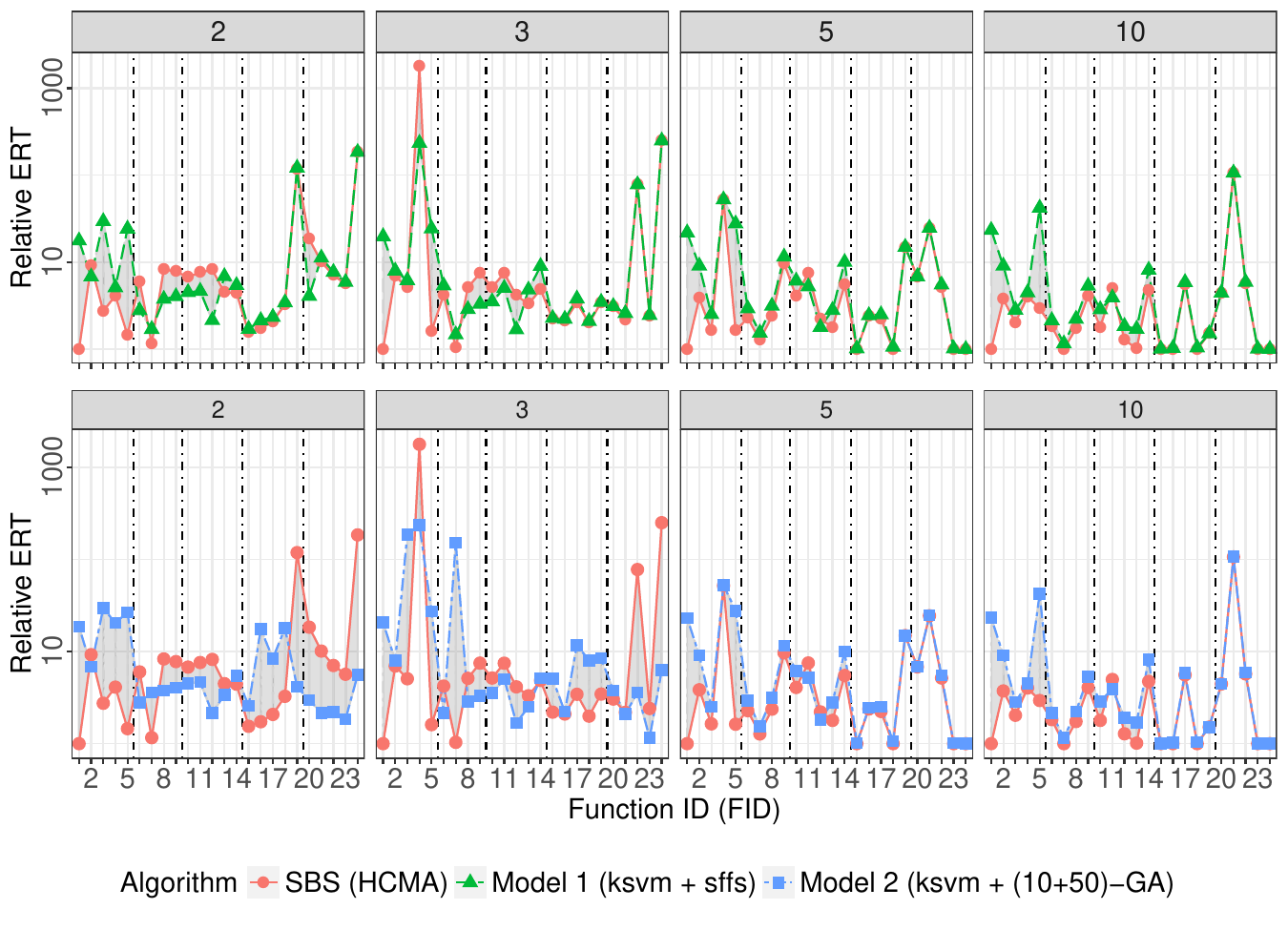}
\caption{Comparison of the relERT (shown on a log-scale) of the single best solver (HCMA, depicted as \textcolor{myred}{$\bullet$}), and the best two algorithm selectors: a kernel-based SVM whose features were selected using a greedy forward-backward strategy (Model 1, \textcolor{mygreen}{$\blacktriangle$}) and an extension of the previous model, which performed a $(10+50)$-GA feature selection on top (Model 2, \textcolor{myblue}{$\blacksquare$}). The performances are shown separately per dimension and selector (top: Model 1, bottom: Model 2). For better visibility, the areas between the curves are highlighted grey and the five BBOB groups are separated from each other by vertical dot-dashed lines.}
\label{fig:sbs_mod_all_pcp}
\end{figure}

These findings are also visible in the upper row of Figure~\ref{fig:sbs_mod_all_pcp}, which compares the relative ERTs of HCMA and Model 1 (on a log-scale) across all 96 BBOB problems (distinguished by FID and dimension). For better visibility of the performance differences, the area between the relative ERTs of the SBS (depicted by \textcolor{myred}{$\bullet$}) and Model 1 (\textcolor{mygreen}{$\blacktriangle$}) is highlighted in grey. One major contribution of our selector obviously is to successfully avoid using HCMA on the three-dimensional version of FID 4. Moreover, one can see that the SBS mainly dominates our selector on the separable BBOB problems (FIDs 1 to 5), whereas our selector achieved equal or better performances on the majority of the remaining problems.

\subsection{Improving the Currently Best Algorithm Selector}
\label{sec:model2}

Although Model 1 already performs better than the SBS, we tried to improve the previous model by making use of results from previous works. More precisely, we start with the set of the eight features that were employed by Model 1 and try to find a more informative feature subset by adding the meta-model and nearest better clustering features. The rational behind this approach is that (a) these feature sets -- as shown in \cite{Kerschke2015} -- are very useful for characterizing a landscape's global structure, and (b) the feature set of Model 1 was derived from a greedy approach and therefore might have missed better feature subsets.

Based on the combination of the three feature sets from above (i.e., the features from Model 1, extended by the NBC and meta-model feature sets), additional feature selections were performed (using the four feature selection strategies described in Section~\ref{sec:featsel}). Two of the new selectors, namely the ones constructed using the greedy sfbs-approach (with a mean relative ERT of 14.27) and the $(10+50)$-GA (mean relative ERT: 14.24) -- performed better than Model 1. The latter model, i.e., the one based on the genetic algorithm, is from here on denoted as ``Model 2''.

As displayed in Table~\ref{tab:portfolio}, Model 2  not only has the best overall performance, but also performs best on the multimodal functions with a weak global structure (FIDs 20 to 24). This effect is also visible in the right column of Figure~\ref{fig:sbs_mod_all}, where the problems of the respective BBOB group (colored in \textcolor{mypink}{pink}) are located either in the left half of the respective plots (for 2D and 3D) or at least on the diagonal itself (5D and 10D).

Also, Figures~\ref{fig:sbs_mod_all} and~\ref{fig:sbs_mod_all_pcp} reveal that Model 2 differs more often from the SBS than Model 1. This finding is also supported by Table~\ref{tab:preds}, which summarizes the actual and predicted solvers. More precisely, it lists (in the left column) how often each of the 12 optimization algorithms from the portfolio performed best. The remaining columns list the predicted optimization algorithms. For instance, the first row of the table shows that for six problems, BSqi had the best performance among all 12 solvers from the portfolio. However, instead of predicting BSqi, Model 1 predicted fmincon and HCMA three times each, and Model 2 predicted fmincon, HCMA and HMLS (each of them twice) for these six problems. Interestingly, one can observe that -- although HCMA only in roughly 15\% of the problems (14 of 96) is the true best solver -- both models show a strong bias towards the SBS: Model~1 selects it in approximately 80\% and Model~2 in 46\% of the problems. Another remarkable observation is that the two models only predict three and four different solvers, respectively.

The different behavior of the two selectors is likely caused by the differences in the used features. While three of the features are identical for both selectors -- the previously described angle and levelset features, as well as the skewness of the objective values -- the remaining six features of Model 2 are meta-model and nearest better clustering (NBC) features. The two meta-model features measure (1) the smallest absolute, non-intercept coefficient of a linear model (without interactions) and (2) the adjusted model fit ($R_{adj}^2$) of a quadratic model \citep[without interactions; see][]{Mersmann2011}. The remaining four NBC features measure (1) the ratio of the standard deviations of the distances among the nearest neighbours and the nearest better neighbours, (2) the ratio of the respective arithmetic means, (3) the correlation between the distances of the nearest neighbours and the nearest better neighbours, and (4) the so-called \textit{indegree}, i.e., the correlation between fitness value and the count of observations to whom the current observation is the nearest better neighbour \citep{Kerschke2015}.

\begin{table}[t]
\caption{Comparison of the predicted solvers. Each row shows how often the respective best solver was predicted as fmincon, HCMA, HMLSL or MLSL by the selectors (Model 1 / Model 2).}
\label{tab:preds}
\centering
\begin{tabular}{rr|rrrr}
  \toprule
  \multicolumn{2}{c|}{True Best Solver} & \multicolumn{4}{c}{Predicted Solver (Model 1 / Model 2)} \\
  \cmidrule{3-6}
 Solver & \# & fmincon & HCMA & HMLSL & MLSL \\ 
  \midrule
  BSqi & 6 &  3 /  \phantom{0}2 &  3 /  \phantom{0}2 &  0 /  \phantom{0}2 &  0 /  \phantom{0}0 \\ 
  BSrr & 6 &  1 /  \phantom{0}2 &  5 / \phantom{0}4 &  0 /  \phantom{0}0 &  0 /  \phantom{0}0 \\ 
  CMA-CSA & 7 &  0 /  \phantom{0}1 &  7 / \phantom{0}3 &  0 /  \phantom{0}3 &  0 /  \phantom{0}0 \\ 
  fmincon & 12 &  0 /  \phantom{0}4 &  8 / \phantom{0}4 &  0 /  \phantom{0}1 &  4 /  \phantom{0}3 \\ 
  fminunc & 6 &  1 /  \phantom{0}2 &  4 / \phantom{0}3 &  0 /  \phantom{0}0 &  1 /  \phantom{0}1 \\ 
  HCMA & 14 &  0 /  \phantom{0}3 & 14 / 11 &  0 /  \phantom{0}0 &  0 /  \phantom{0}0 \\ 
  HMLSL & 11 &  3 /  \phantom{0}3 &  7 /  \phantom{0}4 &  0 /  \phantom{0}0 &  1 /  \phantom{0}4 \\ 
  IPOP400D & 7 &  0 /  \phantom{0}0 &  7 / \phantom{0}3 &  0 /  \phantom{0}3 &  0 /  \phantom{0}1 \\ 
  MCS & 4 &  2 /  \phantom{0}1 &  2 / \phantom{0}0 &  0 /  \phantom{0}3 &  0 /  \phantom{0}0 \\ 
  MLSL & 12 &  4 /  \phantom{0}2 &  8 / \phantom{0}6 &  0 /  \phantom{0}2 &  0 /  \phantom{0}2 \\ 
  OQNLP & 6 &  0 /  \phantom{0}1 &  6 / \phantom{0}2 &  0 /  \phantom{0}2 &  0 /  \phantom{0}1 \\ 
  SMAC-BBOB & 5 &  0 /  \phantom{0}0 &  5 / \phantom{0}2 &  0 /  \phantom{0}1 &  0 /  \phantom{0}2 \\ 
  \midrule
  $\Sigma$ & 96 & 14 / 21 & 76 / 44 & 0 / 17 & 6 / 14 \\
  \bottomrule
\end{tabular}
\end{table}


\subsection{Critical Discussion: Robustness of the Results}

Based on the previous results, one might argue that the proposed approach only performs that well due to HCMA's problem on the three-dimensional version of the \textit{B\"uche-Rastrigin} function (FID 4). Therefore, we repeated our experiments with a subset of all 95 BBOB problems leaving out the problematic instance. Following the approach that we already employed for constructing Model 2 (Section~\ref{sec:model2}), we again found algorithm selection models that outperform HCMA on this subset. The best of them is a classification-based random forest using $y$-distribution (2), meta-model (5), basic, cell mapping, information content and nearest better clustering (1 each) features. As shown in Table~\ref{tab:noFID4}, the random forest outperforms HCMA -- despite the feature costs -- in each of the dimensions and requires on average less than two thirds of the function evaluations.

Moreover, analyzing Table~\ref{tab:portfolio}, one might eventually draw the conclusion that the optimization algorithms mainly perform differently on lower dimensional problems (2D and 3D) and that our presented approach thus might not be effective enough on higher dimensional problems. One could even argue that HCMA outperforms Models 1 and 2 (Sections~\ref{sec:model1} and \ref{sec:model2}) on the latter ones. Although the results in Table~\ref{tab:noFID4} already disprove this hypothesis, we additionally repeated our experiments based on the 48 higher dimensional BBOB problems (5D and 10D) only to support the effectiveness of our approach.

Again, HCMA showed the best performance among all solvers within the portfolio, having very competitive relative ERTs of 6.5 (5D) and 6.9 (10D), respectively. Nevertheless, we trained new algorithm selection models and again found models that substantially outperform HCMA on this particular subset. The best of them is a classification-based random forest using two $y$-distribution, six meta-model, two information content and two nearest better clustering features.

Table~\ref{tab:5and10} compares the performances of HCMA and the random forest per group of BBOB problems. For most of them, the relative ERTs of our algorithm selector (including feature costs) are below the ones of the SBS and in consequence the random forest also had a lower mean relative ERT (4.5), indicating that it (on average) requires only two thirds of the number of function evaluations compared to HCMA.

\begin{table}[t]
\begin{minipage}[t]{0.435\textwidth} 
\caption{Per problem dimension aggregated relative ERTs of the SBS and best algorithm selector (a classification-based random forest), when ignoring the (for HCMA problematic) three-dimensional version of the \textit{B\"uche-Rastrigin} function (FID 4).}
\label{tab:noFID4}
\centering
\begin{tabular}{c|rr}
   \toprule  Problem & \multirow{2}{*}{SBS} & \multirow{2}{*}{AS} \\ 
   Dimension & & \\
   \midrule  \phantom{0}2D & 17.7 & 17.2 \\ 
   \phantom{0}3D & 17.6 & 5.1 \\ 
   \phantom{0}5D & 6.5 & 4.7 \\ 
  10D & 6.9 & 4.5 \\ 
    \midrule all & 12.1 & 7.9 \\ 
    \bottomrule
\end{tabular}
\end{minipage}
\hfill
\begin{minipage}[t]{0.535\textwidth} 
\caption{Comparison of the performances of the single best solver (SBS), i.e., HCMA, and the best algorithm selector (AS), i.e., a classification-based random forest using 12 ELA features, on a subset of the 48 higher dimensional (5 and 10D) BBOB problems.}
\label{tab:5and10}
\centering
\begin{tabular}{c|rr|rr|rr}
   \toprule \!BBOB-\! & \multicolumn{2}{c|}{\!5D\!} & \multicolumn{2}{c|}{\!10D\!} & \multicolumn{2}{c}{\!all\!} \\
     \!Group\! & \!SBS\! & \!AS\! & \!SBS\! & \!AS\! & \!SBS\! & \!AS\! \\ 
   \midrule \!F1 - F5\! & \!12.0\! & \!11.7\! & \!2.7\! & \!14.8\! & \!7.4\! & \!13.3\! \\ 
  \!F6 - F9\! & \!3.9\! & \!2.1\! & \!2.2\! & \!1.6\! & \!3.0\! & \!1.9\! \\ 
  \!F10 - F14\! & \!4.2\! & \!3.7\! & \!2.8\! & \!2.8\! & \!3.5\! & \!3.2\! \\ 
  \!F15 - F19\! & \!4.3\! & \!3.1\! & \!2.0\! & \!2.1\! & \!3.2\! & \!2.6\! \\ 
  \!F20 - F24\! & \!7.7\! & \!1.2\! & \!23.6\! & \!1.3\! & \!15.7 & \!1.3\! \\ 
    \midrule \!all\! & \!6.5\! & \!4.4\! & \!6.9\! & \!4.6\! & \!6.7\! & \!4.5\! \\ 
    \bottomrule \end{tabular}
\end{minipage}
\end{table}

\section{Conclusion and Outlook}
\label{sec:conclusion}

We showed that sophisticated machine learning techniques combined with informative exploratory landscape features form a powerful tool for automatically constructing algorithm selection models for unseen optimization problems. This work provides the first extensive study on feature-based algorithm selection in case of single-objective continuous black-box optimization. With our approach, we were able to reduce the mean relative ERT of the single best solver of our portfolio by half relying only on a quite small set of ELA features and function evaluations. A specific \texttt{R}-package enhanced by a user-friendly graphical user interface allows for transferring the presented methodology to other (continuous) optimization scenarios differing from the BBOB setting.

Of course, the quality and usability of such models heavily relies on the quality of the algorithm benchmark, the considered performance measure, as well as on the representativeness of the included optimization problems. While we considered the well-known and commonly accepted BBOB workshop setting, we are aware of possible shortcomings and will in future work extend our research in terms of including other comprehensive benchmarks and practical applications once available.

Moreover, within this work our experimental studies relied on the ERT, i.e., the most commonly used performance measure for single-objective continuous optimization. However, despite the ERT's current status as gold standard within this domain, its unquestioned usage across any possible setting is at least debatable. In fact, other, also multi-objective, performance measures \citep[see, e.g.,][]{vanrijn2017,bossek2018,kerschke2018gecco} might be more applicable. And although our approach is easily transferrable to other settings (including different performance measures), changing the underlying metric likely results in a different algorithm selection model. Within future research, we thus intend to investigate alternative performance measures in detail and assess their applicability for algorithm selection purposes.

In addition, we also intend to investigate the potential of more sophisticated feature selection strategies, as well as of automatically tuning the hyperparameters of the included machine learning techniques. Finally, we work on extending the methodological approach to constrained and multi-objective optimization problems \citep{KerschkeWPGDTE2016,KerschkeG2017,Kerschke2018} in order to increase the applicability to real-world scenarios.


\section*{Acknowledgment}

The authors acknowledge support from the \textit{European Research Center for Information Systems\footnote{\url{https://www.ercis.org/}} (ERCIS)} and the DAAD PPP project No. 57314626.

\bibliographystyle{apalike}
\bibliography{ecj}

\end{document}